%% file: main.tex
\documentclass[10pt,twocolumn,letterpaper]{article}
\usepackage[final]{cvpr}

\usepackage{graphicx}
\usepackage{amsmath}
\usepackage{amssymb}
\usepackage{booktabs}

\usepackage[accsupp]{axessibility}  % Improves PDF readability for those with disabilities.

\usepackage[pagebackref,breaklinks,colorlinks,bookmarks=false]{hyperref}

\usepackage{multirow}
\usepackage{enumitem}
\usepackage{rotating}
\usepackage{makecell}
\usepackage{marvosym}

\usepackage{ragged2e}
\usepackage{caption}
\usepackage{subcaption}
\usepackage[normalem]{ulem}
\usepackage[capitalize]{cleveref}
\crefname{section}{Sec.}{Secs.}
\Crefname{section}{Section}{Sections}
\Crefname{table}{Table}{Tables}
\crefname{table}{Tab.}{Tabs.}
\usepackage{color}
\usepackage{xcolor}
\usepackage{color, colortbl}
\usepackage{booktabs}
\usepackage{float}
\usepackage{wrapfig}
\definecolor{Gray}{gray}{0.9}
\definecolor{Red}{RGB}{230, 57, 70}
\definecolor{Blue}{RGB}{0, 100, 148}
\usepackage{mmstyle}
\usepackage{xspace}

\newcommand{\modelname}{\emph{OSX}\xspace}
\newcommand{\dataname}{\emph{UBody}\xspace}

\newcommand\blfootnote[1]{%
  \begingroup
  \renewcommand\thefootnote{}\footnote{#1}%
  \addtocounter{footnote}{-1}%
  \endgroup
}

\def\cvprPaperID{3184} % *** Enter the CVPR Paper ID here
\def\confName{CVPR}
\def\confYear{2023}
\begin{document}
\title{One-Stage 3D Whole-Body Mesh Recovery with Component Aware Transformer}  % **** Enter the paper title here

\author{%
	Jing Lin$^{1,2\S}$, Ailing Zeng$^{1\P}$, Haoqian Wang$^{2}$, Lei Zhang$^{1}$, Yu Li$^{1}$ \\
		$^{1}$ International Digital Economy Academy (IDEA), \\ $^2$  Shenzhen International Graduate School, Tsinghua University \\
		\url{https://osx-ubody.github.io}
}
%\maketitle 

\vspace{-0.75cm}
\twocolumn[{%
    \renewcommand\twocolumn[1][]{#1}%
    \maketitle
    \centering
    \captionsetup{type=figure}
    \includegraphics[width=0.9\linewidth]{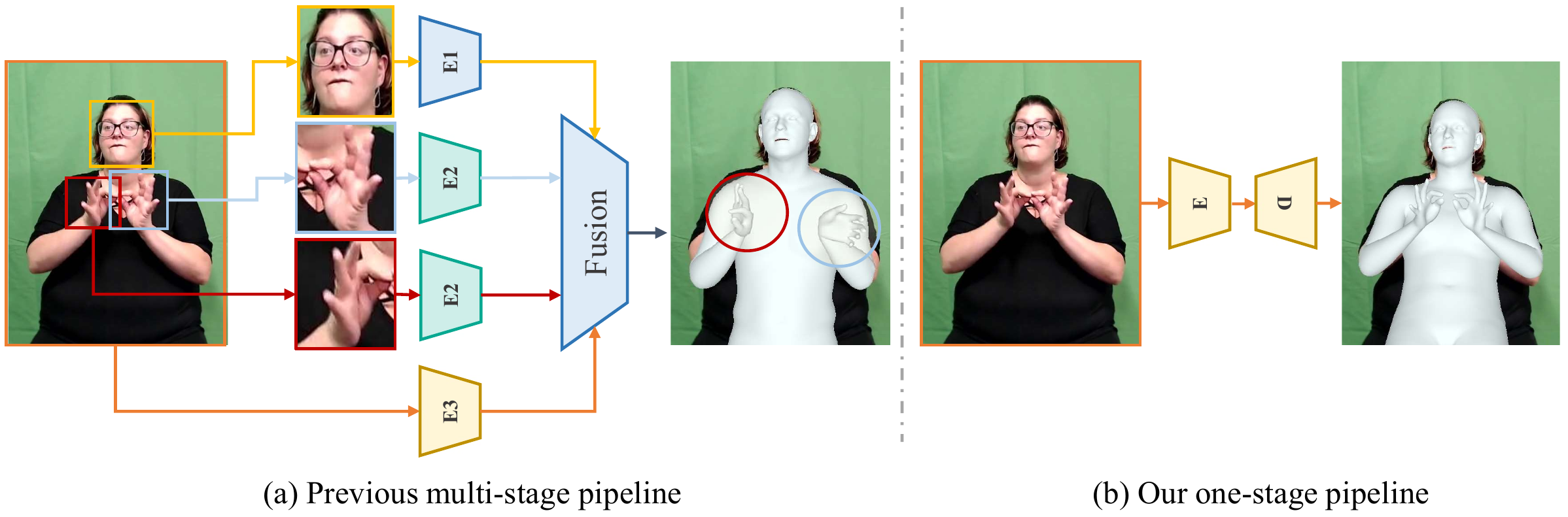}
    % \vspace{-0.2cm}
    \caption{A comparison of existing whole-body mesh recovery methods and ours. Most existing methods leverage a multi-stage pipeline which uses separate expert models to process body component (\eg, \textbf{E1}: HeadNet, \textbf{E2}: HandNet, \textbf{E3}: BodyNet) and fuse them to get the whole-body prediction in a copy-paste manner. The result (from~\cite{PavlakosGeorgios2020expose}) produces unnatural wrist poses. In contrast, our pipeline is a neat one-stage framework with a single encoder-decoder and can predict more accurately with natural meshes.}
    \vspace{0.7cm}
    \label{fig:teaser}
}]
\vspace{0.7cm}

%%%%%%%%% ABSTRACT
\begin{abstract}

%  final
Whole-body mesh recovery aims to estimate the 3D human body, face, and hands parameters from a single image. It is challenging to perform this task with a single network due to resolution issues, i.e., the face and hands are usually located in extremely small regions. Existing works usually detect hands and faces, enlarge their resolution to feed in a specific network to predict the parameter, and finally fuse the results. While this copy-paste pipeline can capture the fine-grained details of the face and hands, the connections between different parts cannot be easily recovered in late fusion, leading to implausible 3D rotation and unnatural pose. In this work, we propose a one-stage pipeline for expressive whole-body mesh recovery, named OSX, without separate networks for each part. Specifically, we design a Component Aware Transformer (CAT) composed of a global body encoder and a local face/hand decoder. The encoder predicts the body parameters and provides a high-quality feature map for the decoder, which performs a feature-level upsample-crop scheme to extract high-resolution part-specific features and adopt keypoint-guided deformable attention to estimate hand and face precisely. The whole pipeline is simple yet effective without any manual post-processing and naturally avoids implausible prediction. Comprehensive experiments demonstrate the effectiveness of OSX. Lastly, we build a large-scale Upper-Body dataset (UBody) with high-quality 2D and 3D whole-body annotations. It contains persons with partially visible bodies in diverse real-life scenarios to bridge the gap between the basic task and downstream applications. 
%\footnote{The code and dataset will be released upon acceptance.\\
\vspace{0pt}
\blfootnote{$\S$ Work done during an internship at IDEA; ${\P}$~Corresponding author.}
\end{abstract}

\input{sec/intro}
\input{sec/related_work}

\input{sec/method}

\input{sec/data}

\input{sec/exp}

\input{sec/conclusion}
% {\small
% \bibliographystyle{ieee_fullname}
% \bibliography{egbib}
% }

\input{main.bbl}
{\small
\bibliographystyle{ieee_fullname}
}
\input{sec_sup/supp}

%%%%%%%%% REFERENCES

\end{document}

%% file: sec/intro.tex
\vspace{-0.5cm}
\section{Introduction}
Expressive whole-body mesh recovery aims to jointly estimate the 3D human body poses, hand gestures, and facial expressions from monocular images. It is gaining increasing attention due to recent advancements in whole-body parametric models (\eg, SMPL-X~\cite{Pavlakos_2019smplx}). 
This task is a key step in modeling human behaviors and has many applications, \eg, motion capture, human-computer interaction. Previous research focus on individual tasks of reconstructing human body~\cite{Kanazawa_2018_hmr,Kolotouros_2019_spin,zeng2022deciwatch,zeng2022smoothnet,Choi_2020_pose2mesh,tian2022survey}, face~\cite{OswaldAldrian2013InverseRO,AyushTewari2017MoFAMD,YuDeng2019Accurate3F,BernhardEgger20203DMF}, or hand~\cite{AdnaneBoukhayma20193DHS,LinHuang2021SurveyOD,TheocharisChatzis2020ACS}. However, whole body mesh recovery is particularly challenging as it requires accurate estimation of each part and natural connections between them.

Existing learning-based works~\cite{PavlakosGeorgios2020expose,Feng_2021_pixie,Rong_2021frank,GyeongsikMoon2020hand4whole,HongwenZhang2022PyMAFXTW} use multi-stage pipelines for body, hand, and face estimation to achieve the goal of this task. 
As depicted in Figure~\ref{fig:teaser}(a), these methods typically detect different body parts, crop and resize each region, and feed them into separate expert models to estimate the parameters of each part. 
The multi-stage pipeline with different estimators for body, hand, and face results in a complicated system with a large computational complexity. Moreover, the blocked communications among different components inevitably cause incompatible configurations, unnatural articulation of the mesh, and implausible 3D wrist rotations as they cannot obtain informative and consistent clues from other components. 
Some methods~\cite{Feng_2021_pixie,GyeongsikMoon2020hand4whole,HongwenZhang2022PyMAFXTW} attempt to alleviate these issues by designing additional complicated integration schemes or elbow-twist compensation fusion among individual body parts. However, these approaches can be regarded as a late fusion strategy and thus have limited ability to enhance each other and correct implausible predictions. 

In this work, we propose a one-stage framework named \modelname for 3D whole-body mesh recovery, as shown in Figure~\ref{fig:teaser}(b), which does not require separate networks for each part.
Inspired by recent advancements in Vision Transformers~\cite{dosovitskiy2020vit,YufeiXu2022ViTPoseSV}, which are effective in capturing spatial information in a plain architecture, we design our pipeline as a component-aware Transformer (CAT) composed of a global body encoder and a local component-specific decoder. The encoder equipped with body tokens as inputs captures the global correlation, predicts the body parameters, and simultaneously provides high-quality feature map for the decoder.
The decoder utilizes a differentiable upsample-crop scheme to extract part-specific high-resolution features and adopt the keypoint-guided deformable attention to precisely locate and estimate hand and face parameters. 
 The proposed pipeline is simple yet effective without any manual post-processing. To the best of our knowledge, this is the first one-stage pipeline for 3D whole-body estimation. 
We conduct comprehensive experiments to investigate the effects of the above designs and compare our method, with existing works on three benchmarks. Results show that \modelname outperforms the state-of-the-art (SOTA)~\cite{GyeongsikMoon2020hand4whole} by $9.5$\% on AGORA, $7.8$\% on EHF, and $13.4$\% on the body-only 3DPW dataset.

\begin{figure*}[ht]
\begin{center}
\includegraphics[width=0.98\linewidth]{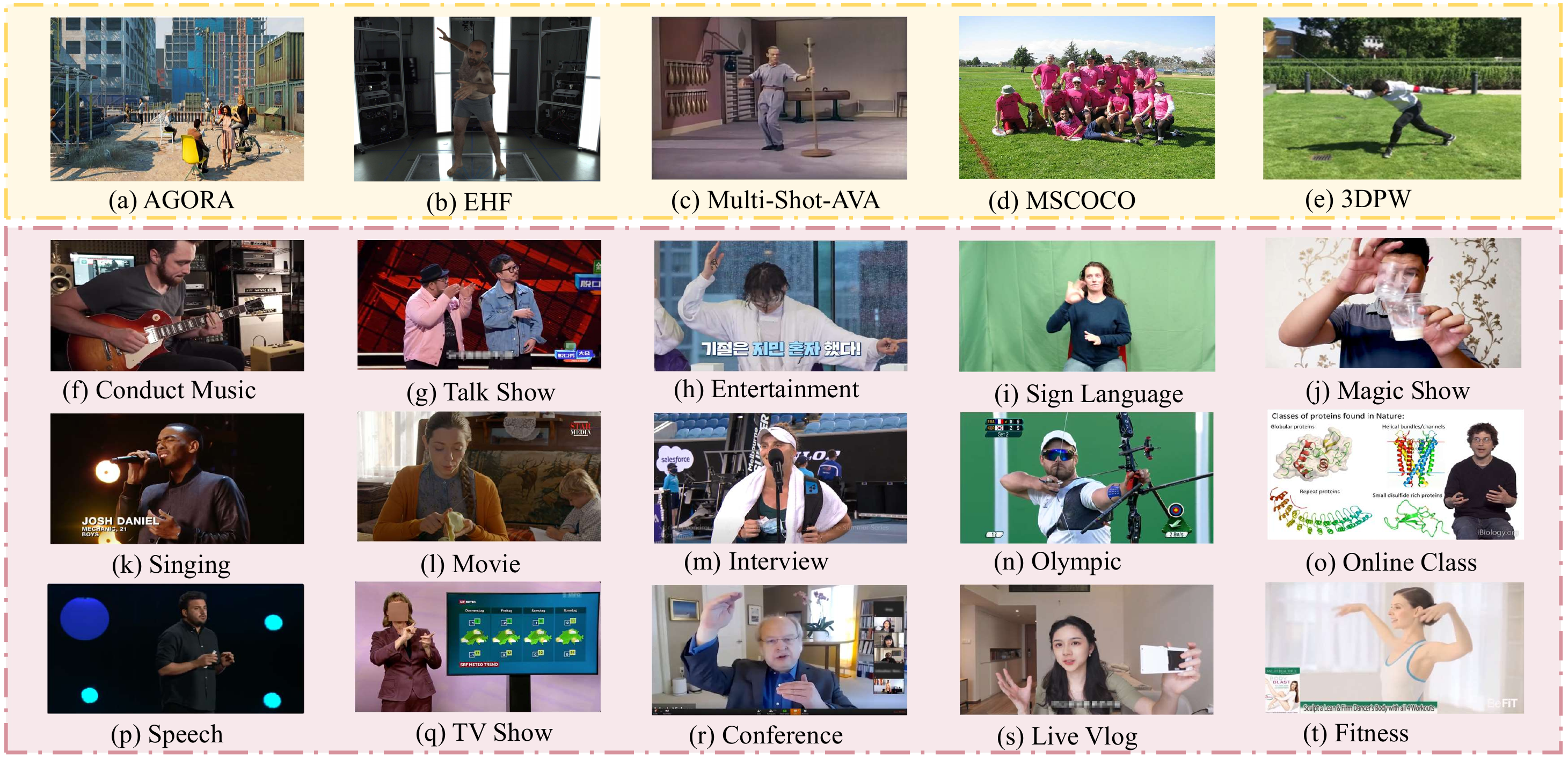}
\end{center}
\vspace{-0.6cm}
\caption{
Illustration of five previous datasets (from (a) to (e)) and the proposed Upper Body Dataset (from (f) to (t)) with fifteen real-life scenes. \dataname bridges the gap between the basic 3D whole-body estimation task and downstream tasks with highly expressive actions.
}
\label{fig:ubody}
\vspace{-0.6cm}
\end{figure*}

In addition, existing popular benchmarks, as illustrated in the first row of Figure~\ref{fig:ubody}, are either indoor single-person scenes with limited images (\eg, EHF~\cite{Pavlakos_2019smplx}) or outdoor synthetic scenes (e.g., AGORA~\cite{Patel_2021agora}), where the people are often too far from the camera and the hands and faces are frequently obscured. 
In fact, human pose estimation and mesh recovery is a fundamental task that benefits many downstream applications, such as sign language recognition, gesture generation, and human-computer interaction.
Many scenarios, such as talk shows and online classes, are of vital importance to our daily life yet under-explored. In such scenarios, the upper body is a major focus, whereas the hand and face are essential for analysis.
To address this issue, we build a large-scale upper-body dataset with fifteen human-centric real-life scenes, as shown in Figure~\ref{fig:ubody}(f) to (t). This dataset contains many unseen poses, diverse appearances, heavy truncation, interaction, and abrupt shot changes, which are quite different from previous datasets. Accordingly, we design a systematical annotation pipeline and provide precise 2D whole-body keypoint and 3D whole-body mesh annotations. With this dataset, we perform a comprehensive benchmarking of existing whole-body estimators. 

Our contributions can be summarized as follows.
\begin{itemize}
\vspace{-0.1cm}
\item We propose a one-stage pipeline, \modelname, for 3D whole-body mesh recovery, which can regress the SMPL-X parameters in a simple yet effective manner.
\vspace{-0.2cm}
\item Despite the conceptual simplicity of our one-stage framework, it achieves the new state of the art on three popular benchmarks.
%We demonstrate the effectiveness of \modelname with 
\vspace{-0.2cm}
\item We build a large-scale upper-body dataset, \dataname, to bridge the gap between the basic task and downstream applications and provide precise annotations, with which we conduct benchmarking of existing methods. We hope \dataname can inspire new research topics. 
\end{itemize}

%% file: sec/related_work.tex
\vspace{-0.2cm}
\section{Related Work}
\label{sec:related}
\subsection{Methods of Whole-body Mesh Recovery}
\label{sec:related_method}
Whole-body mesh recovery targets to localize mesh vertices of all human components, including body, hands, and face from monocular images. Most previous works focus only on individual hand~\cite{AdnaneBoukhayma20193DHS,LinHuang2021SurveyOD,TheocharisChatzis2020ACS}, face~\cite{OswaldAldrian2013InverseRO,AyushTewari2017MoFAMD,YuDeng2019Accurate3F,BernhardEgger20203DMF}, and body~\cite{Kolotouros_2019_spin,zeng2022deciwatch,tian2022survey, kocabas2021pare, kocabas2021spec} reconstruction.
In contrast, the joint whole-body estimation methods are less addressed. %
Some optimization-based works reconstruct 3D bodies by fitting the detected 2D keypoints from images with additional constraints, but they are slow and prone to local optima~\cite{Pavlakos_2019smplx,Xiang_2019_wholebody3d}.
Thanks to the whole-body parametric model (\eg, SMPL-X~\cite{Pavlakos_2019smplx}), learning-based models~\cite{PavlakosGeorgios2020expose,Rong_2021frank,Zhou_2021_full,Feng_2021_pixie, sun2022learning} emerge to train networks to predict expressive body pose, shape, hand gesture, and facial expression. 
Due to the low resolution of hands and face, these whole-body methods crop and resize the hands and face images to higher resolutions and feed them into separate expert networks to conduct the corresponding parameter regression. 
Specifically, ExPose~\cite{PavlakosGeorgios2020expose} introduces body-driven attention for higher-resolution crops of the face and hand estimation, a dedicated refinement module, and part-specific knowledge from existing hand-only and face-only datasets.
FrankMocap~\cite{Rong_2021frank} presents a regression-and-integration method to build a fast and accurate system.
PIXIE~\cite{Feng_2021_pixie} produces animatable whole body with realistic facial details via a moderator to fuse body part features adaptively.
Recently, Hand4Whole~\cite{GyeongsikMoon2020hand4whole} utilizes both body and hand joint features for accurate 3D wrist rotation and smooth connection between body and hands.

Nevertheless, these methods aim at high performance by using separate networks in a divide-and-conquer fashion for different components and a specific fusion module to paste them together.
The multi-stage pipelines lead to high complexity and inevitably cause inconsistent and unnatural articulation of the mesh and implausible 3D wrist rotations, especially in occluded, truncated, and blurry contexts. Until now, one-stage methods in this task are unexplored.

\subsection{Benchmarks of Expressive Body}

Some datasets with parametric model annotations~\cite{Patel_2021agora,Pavlakos_2019smplx,von_Marcard_3dpw,cai2022humman,Ionescu_2014_hm36,Moon_2022NeuralAnnot,HanbyulJoo2022eft} have been developed to advance the field. 
Table~\ref{tab:datasets} summarizes these datasets from the annotation type, size, scene diversity, \etc. 
To be specific, EHF~\cite{Pavlakos_2019smplx} is the first evaluation dataset for SMPL-X-based models, which is built by capturing 3D body shapes with a scanning system and then fitting the SMPL-X model to the scans.
AGORA~\cite{Patel_2021agora} is a synthetic dataset with high realism and accurate ground truth, which is by far the most commonly used test data due to the diversity of subjects, environments, clothes, and occlusions.
Notably, people in AGORA are often far from the camera, and their hands and face are obscured and have small resolutions, making existing methods focus more on body rather than hand and face estimation. 

Since marker-based 3D mocap labels are hard to obtain, there are a few annotation methods~\cite{Moon_2022NeuralAnnot,Pavlakos_2019smplx,Feng_2021_pixie,Muller_2021contact,rockwell2020full,pavlakos2022multishot} for high-precision labeling for both monocular indoor and outdoor scenes.
FBA~\cite{rockwell2020full} emphasizes the severe failure cases of existing body recovery methods on consumer video data due to unusual camera viewpoints and aggressive truncations. They annotate pseudo 2D body keypoints and SMPL annotations via HMR~\cite{Kanazawa_2018_hmr} on 13k frames across four action recognition datasets.
Multi-shot-AVA~\cite{pavlakos2022multishot} also argues that data from edited media, like movies with rich appearances,  interactions between humans, and various temporal contexts, is valuable. They apply the proposed multi-shot optimization on AVA~\cite{ChunhuiGu2017AVAAV} to get pseudo 3D ground truth.
Interestingly, a body recovery benchmark~\cite{pang2022benchmarking} finds that simply using the 2D COCO dataset with pseudo-3D labels can surprisingly achieve a better performance and generalization ability.
To complement these prior datasets and focus on expressive body recovery, we construct a new benchmark with high-quality 2D and 3D whole-body annotations.

\begin{table*}[h]
  \centering
  \resizebox{\textwidth}{!}{
% Table generated by Excel2LaTeX from sheet 'Datasets (2)'
\begin{tabular}{llcccccccc}
\toprule
\textbf{Type} & \multicolumn{1}{l}{\textbf{Dataset}} & \multicolumn{1}{c}{\textbf{\makecell{\#Frames}}} & \multicolumn{1}{c}{\textbf{\makecell{Scenes}}} &  \multicolumn{1}{c}{\textbf{\makecell{~Multi\\Person~}}} & \textbf{\makecell{~In-the-\\wild~}} &\textbf{\makecell{~Upper\\Body~}}&\multicolumn{1}{c}{\textbf{\makecell{~Video~}}} & \textbf{\makecell{Annotation\\Type}} & \textbf{\makecell{Annotation\\Source}} \\
\midrule
\multirow{1}[1]{*}{\makecell{Rendered}} 
      & AGORA~\cite{Patel_2021agora} & 17K   &Daily    & Y & N   &N &N & SMPL-X & \cite{Patel_2021agora} \\
\midrule
\multirow{2}[2]{*}{\makecell{Marker/Sensor-\\based MoCap}} 
      & Human3.6M~\cite{Ionescu_2014_hm36} & 3.6M  & Daily      & N     &N   &N  &Y    & SMPL-X  & \cite{Moon_2022NeuralAnnot} \\
      &  3DPW~\cite{von_Marcard_3dpw} &  $>$ 51K &  Daily &  Y &   Y&N & Y& SMPL-X &  \cite{Moon_2022NeuralAnnot} \\
\midrule
\multirow{3}[2]{*}{\makecell{Marker-less\\Multi-view\\MoCap}} 
      &  MPI-INF-3DHP~\cite{mehta2017monocular} &  $>$ 1.3M &  Daily &   N &  Y&N & Y& SMPL-X &  \cite{Moon_2022NeuralAnnot} \\
      & EHF~\cite{Pavlakos_2019smplx} & 0.1K   & Daily        & N     & N   &N    &N& SMPL-X & \cite{Pavlakos_2019smplx} \\
      & ZJU-MoCap~\cite{peng2021neural} &  $\geq$ 237K     & Daily     & N     & N   &N&Y     & SMPL-X & \cite{easymocap} \\
\midrule
\multirow{8}[2]{*}{\makecell{\\Pseudo-\\3D Labels}} 
      &  PennAction~\cite{zhang2013penn} &  77K &  Fitness &   N &   Y&N &Y&  SMPL &  \cite{zhang2019predicting} \\
      & MSCOCO~\cite{lin2014coco} & 200K   & Daily      & Y &  Y&N & N&SMPL-X  & \cite{Moon_2022NeuralAnnot} \\
      & COCO-Wholebody~\cite{jin2020wholebody} & 200K   & Daily  & Y &  Y&N&N & 2D KPT  & \cite{jin2020wholebody} \\
      &  MPII~\cite{andriluka2014mpii} &  25K &  Daily  &  Y &  Y&N &  N&SMPL-X &  \cite{Moon_2022NeuralAnnot} \\
      & MTP~\cite{muller2021self} & 3.8K & Daily   & N     &  Y&N &N& SMPL-X & \cite{muller2021self} \\
      & FBA~\cite{rockwell2020full}&13K&Vlog\&Cook\&Daily&Y& Y&N&Y&SMPL&\cite{rockwell2020full}\\
      &Multi-shot-AVA~\cite{pavlakos2022multishot}&350K&Movie&Y& Y&N&Y&SMPL&\cite{pavlakos2022multishot}\\
      \cmidrule{2-10}
      & \textbf{\dataname (Ours)} &$>$1051K & Real-life Scenes  & Y      &  Y&Y &Y& SMPL-X\&2D KPT & Ours \\
    \bottomrule
    \end{tabular}%
     }
     \vspace{-0.1cm}
    \caption{Comparison of related datasets. \dataname is a large-scale upper-body dataset with high-precision whole-body annotations. }
    \vspace{-0.2cm}
  \label{tab:datasets}%
\end{table*}%

%% file: sec/method.tex
\section{Method}
\label{sec:method}

\begin{table}
\centering
\scalebox{0.85}{
\begin{tabular}{l|ccc|ccc}
\toprule
\multirow{2}{*}{\textbf{Method}}& \multicolumn{3}{c|}{\textbf{AGORA-val}}  & \multicolumn{3}{c}{\textbf{EHF}}      \\
\cmidrule{2-7}
& \textbf{Hand}  & \textbf{Face} & \textbf{All}   & \textbf{Hand} & \textbf{Face} & \textbf{All}   \\
\midrule
Ori.              & \underline{73.3}  & \underline{81.4} & \underline{183.8} & \underline{42.7} & 25.7 & 77.5  \\
% Ori.+1/2 Hand & 73.7  & 81.6 & 183.3 & 45.2 & 25.7 & 78.0    \\
Ori.+1/4 Hand & 75.7  & \textbf{80.8} & \textbf{183.0}   & 50.9 & 24.8 & 78.5  \\
% Ori.+1/2 Face & \textbf{73.0} & 81.6 & 184.6 & 42.6 & 25.2 & 77.6  \\
Ori.+1/4 Face & \textbf{73.2}  & 81.8 & 184.0   & \textbf{41.3} & \underline{24.2} & \textbf{77.0}    \\
\midrule
Share Backbone    & 81.1  & 91.0   & 202.3 & 55.5 & 33.5 & 84.7  \\
Share+1/4 Hand & 77.4  & 86.1 & 188.6& 57.0   & 25.8 & 84.8  \\
Share+1/4 Face & 79.5  & 85.0   & 196.7 & 57.8 & \textbf{24.4} & 82.5 \\
\bottomrule
\end{tabular}}
\vspace{-0.2cm}
\caption{A preliminary study on the effect of different component scales and share backbone for all components' feature extraction.}
\label{tab:motivation}
\vspace{-0.7cm}
\end{table}

\subsection{Motivation}
\label{sec:motivate}

A one-stage framework is vital to simplify the cumbersome processes without hand-craft and complex integration designs. 
However, translating from multi-stage methods directly to a one-stage method is nontrivial.
We take the present state-of-the-art method Hand4Whole~\cite{GyeongsikMoon2020hand4whole} as an example to perform some preliminary studies on bringing the gap between the multi-stage method and one-stage approach. 
On the one hand, we replace its separate backbones with a shared backbone for all human components. On the other hand, we explore different crop-and-resize image resolutions for the hands and face, as they usually have small image resolutions.

Table~\ref{tab:motivation} shows that, when we transition from the original setup (\emph{Ori.}) to a shared backbone (\emph{Share Backbone}), all recovery errors are severely deteriorated on two datasets. Specifically, MPVPE increases from $183.8$mm to $202.3$mm (a \textbf{10.1}\% drop) on AGORA~\cite{Patel_2021agora}, and from $77.5$mm to $84.7$mm (a \textbf{9.3}\% drop) on EHF for all components (\emph{All}). 
These results indicate that extracting the multi-component whole-body features with a shared backbone is difficult. Notably, the hand estimation performance deteriorates by
\textbf{30.0}\% on EHF.
Based on the results of different resolutions, we summarize some interesting observations as follows: 
(i)~Overall, changing the resolution of the hand results in a larger performance drop than the face on EHF;
(ii)~When not sharing a backbone, the results are generally worse with smaller input resolutions of the hands and face.

\begin{figure*}[t]
\begin{center}
\includegraphics[width=.95\linewidth]{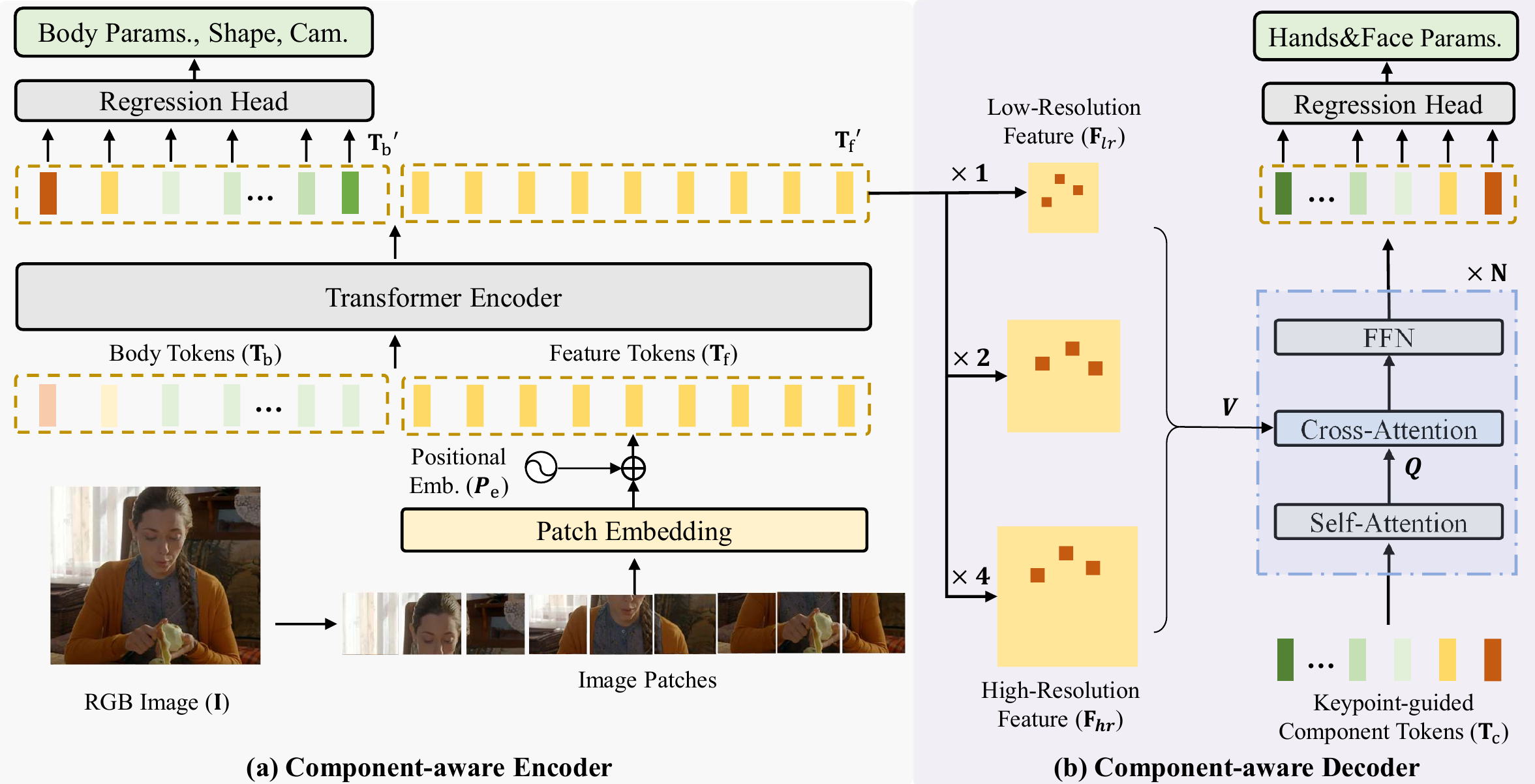}
\end{center}
\vspace{-0.4cm}
\caption{
The overview of the proposed one-stage framework (\modelname) with component-aware transformer. It includes (a) a component-aware Transformer encoder and (b) a component-aware Transformer decoder.
}
\label{fig:method}
\vspace{-0.4cm}
\end{figure*}

\subsection{Building Component Aware Transformer}
\label{sec:method_1}

 As an attempt to break the above status quo, we propose a one-stage framework with a vision transformer encoder and decoder for expressive full-body mesh recovery, named \modelname. It is simple in design and effective in full-body mesh prediction, as we will demonstrate later. We hope it can serve as a baseline for future one-stage methods.
Given a human image $\mI \in \mathbb{R}^{H \times W \times 3}$, our component-aware Transformer (\emph{CAT}) estimates the corresponding body, hand, and face parameters $\hat{\cP}= \{\hat{\mP}_{body}, \hat{\mP}_{lhand}, \hat{\mP}_{rhand}, \hat{\mP}_{face}\}$ and then feed them into a SMPL-X layer~\cite{Pavlakos_2019smplx} to obtain the final 3D whole-body human mesh. 
Specifically, $\hat{\mP}_{body}$ contains 3D body joint rotation $\theta_{body} \in \mathbb{R}^{22 \times 3}$, body shape $\beta\in \mathbb{R}^{10}$, and 3D global translation $t\in \mathbb{R}^{3}$. %$\textbf{t}_{body}\in \mathbb{R}^{3}$. 
For $\hat{\mP}_{lhand}$ and $\hat{\mP}_{rhand}$, they have 3D left and right hand joint rotation $\theta_{lhand} \in \mathbb{R}^{15 \times 3}$ and $\theta_{rhand} \in \mathbb{R}^{15 \times 3}$, respectively. 
$\hat{\mP}_{face}$ consists of 3D jaw rotation $\theta_{face}\in \mathbb{R}^{3}$ and facial expression $\phi \in \mathbb{R}^{10}$.
Our training target is to minimize the distance between the recovered parameters $\hat{\cP}$ and the ground-truth parameters $\cP$. As shown in Figure~\ref{fig:method}, the proposed \emph{CAT} consists of a component-aware encoder to capture the global correlation and extract high-quality multi-scale feature, and a component-aware decoder to strengthen the hand and face regression via an up-sampling strategy to obtain higher-resolution feature maps.

\subsection{Body Regression via Global Encoder}
\label{sec:method_2}
In the component-aware encoder, the human image $\mI$ is split into fixed-size image patches $\mathbf{P}\in \mathbb{R}^{\frac{HW}{M^2}\times (M^2\times3)}$, where $M$ is the patch size. The patches $\mathbf{P}$ are then linearly projected by a convolution layer %with a convolution layer into a feature with the shape of $\frac{H}{d} \times \frac{W}{d} \times C$,%
and added with position embeddings $\mathbf{P_e} \in \mathbb{R}^{\frac{HW}{M^2}\times C}$ to obtain a sequence of feature tokens $\mathbf{T_f}\in \mathbb{R}^{\frac{HW}{M^2}\times C}$.
To explicitly leverage the body prior and learn the body information in the encoder, we concatenate the feature token $\mathbf{T_f}$ with the body tokens $\mathbf{T_b} \in \mathbb{R}^{B  \times C}$, which are learnable parameters. The concatenated tokens are then fed into a standard Transformer encoder with multiple Transformer blocks~\cite{dosovitskiy2020vit}. Each block consists of a multi-head self-attention, a feed-forward network (FFN), and two layer normalization. 
After the global feature fusion, the body tokens and image feature tokens are updated into $\mathbf{T_b}' \in \mathbb{R}^{B \times C}$ and  $\mathbf{T_f}'\in \mathbb{R}^{\frac{HW}{M^2}\times C}$. Finally, we use several fully connected layers to regress the body parameters $\hat{\mP}_{body}=\{\theta_{body} , \beta, t\}$ based on $\mathbf{T_b}'$.

\subsection{High-Resolution Decoder for Hand and Face}
\label{sec:method_3}
% how to obtain precise estimation on low-resolution hand and face estimation
\noindent \textbf{Up-sampling for multi-scale high-resolution features.}
Since the hands and face in a human image are usually small, previous methods upsample the human image and crop out the hands and face to obtain higher-resolution images. However, this image-level upsampling-crop scheme requires additional backbones to extract the hand and face features separately. 
To solve this problem, we propose a differentiable feature-level upsampling-crop strategy to enhance the hands and face regression process as inspired by the recent ViTDet~\cite{Li2022vitdet}.
%
% Specifically, to relieve the cropping quantization error on a low-resolution feature map, 
Specifically, we reshape the feature tokens $\mathbf{T_f}'$ into a feature map and upsample it into multiple higher-resolution features $\mathbf{T}_{hr}$ via deconvolution layers. 
Then, since decoding the hand and face component information from the full feature map inevitably leads to redundant computation and makes the computation process inefficient, we perform differentiable RoIAlign~\cite{He_2017_maskrcnn} on the feature maps and crop out multi-scale hand feature maps $\mathbf{T}_{hand}$ and face feature maps $\mathbf{T}_{face}$, according to the predicted hand and face bounding boxes, which are regressed from  $\mathbf{T_f}'$ using FFNs. 
The up-sampling and decoding processes for hand and face components are the same, and we illustrate the case of hand parameter regression in detail in Figure~\ref{fig:method}(b).
The cropped multi-scale hand features can be represented as $\mathbf{T}_{hand}=\{\mathbf{F}_{lr}, ..., \mathbf{F}_{hr}\}$. The low-resolution feature $\mathbf{F}_{lr} \in \mathbb{R}^{\frac{H'}{M} \times \frac{W'}{M} \times C}$ is cropped from the original low-resolution feature map, where $H'$ and $W'$ are the height and width of hand image patches. $\mathbf{F}_{hr}$ is the highest-resolution feature. The cropped multi-scale features then serve as memory tokens $\mathbf{V}$ for the keypoint-guided component-aware decoder.
To relieve the computational pressure, we reduce the token dimension from $C$ to $C'$ in the component-aware decoder, where $C' = C/2$. 

\begin{figure*}[h]
\centering
\includegraphics[width=0.97\linewidth]{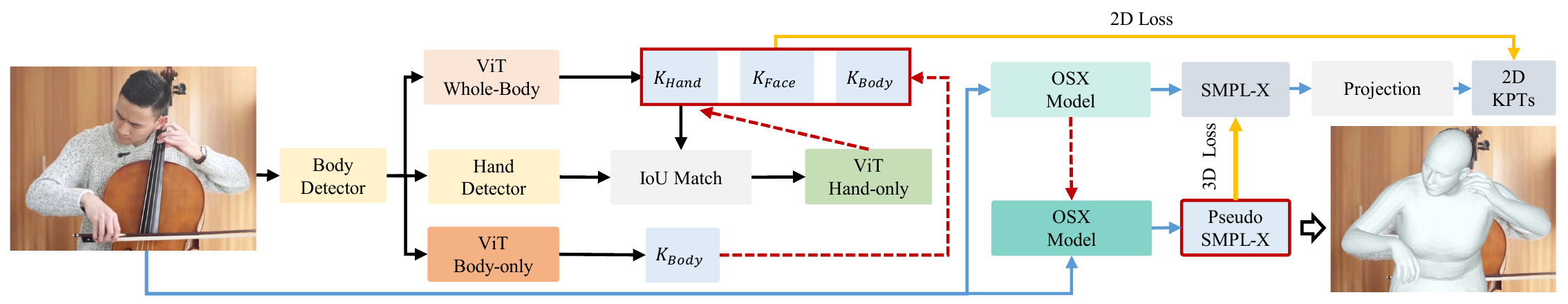}
\vspace{-0.3cm}
\caption{
Illustration of the annotation pipeline of \dataname. Black lines show the annotation process of 2D whole-body keypoints, and blue lines are the 3D SMPL-X annotation procedure. Red dotted lines mean to update the information.
}
\label{fig:ubody_annotation}
\vspace{-0.3cm}
\end{figure*}

\begin{figure*}[h]
\begin{center}
\includegraphics[width=0.97\linewidth]{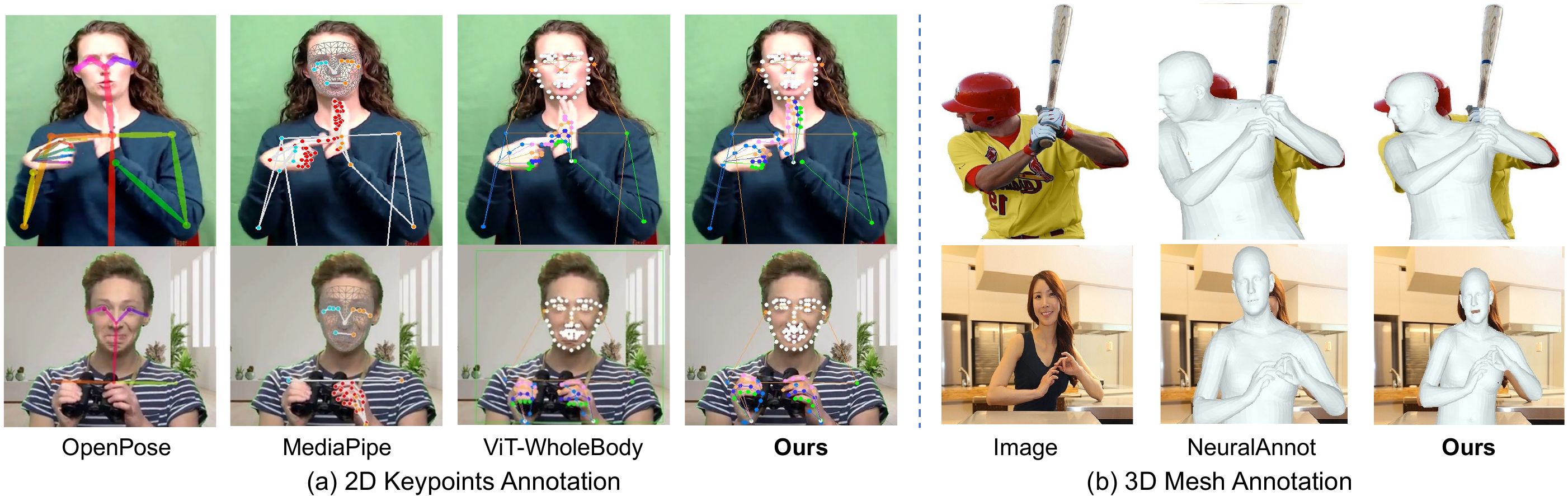}
\end{center}
\vspace{-0.5cm}
\caption{
Comparisons of (a)~the 2D keypoints annotation quality of wildly used methods~\cite{ZheCao2018OpenPoseRM,FanZhang2020MediaPipeHO} and recent SOTA~\cite{YufeiXu2022ViTPoseSV} on \dataname (the left part), and (b)~the 3D mesh annotation quality of previous SOTA~\cite{Moon_2022NeuralAnnot} with ours on COCO (the right part).
}
\label{fig:ubody_vis_2d}
\vspace{-0.1cm}
\end{figure*}
% \vspace{-0.5cm}
\noindent \textbf{Keypoint-guided deformable attention decoder.}
To improve the precision of hand and face parameter regression, we leverage 2D keypoint positions as prior knowledge to obtain better component tokens $\mathbf{T_c}$ than random initialization.
We simply use the feature map $\mathbf{F}_{lr}$ to regress each 2D keypoint to trade off accuracy and efficiency and regard it as a reference keypoint. The input $\mathbf{T_c}\in \mathbb{R}^{K \times C'}$ of the decoder, which we call the keypoint-guided component tokens, is obtained by summing up reference keypoint feature, pose positional embedding, and learnable embeddings.
We then pass the keypoint-guided component token through $N$ deformable attention blocks as inspired by deformable DETR~\cite{Zhu_detr21}. To relieve the issue of looking over all possible spatial locations, these blocks learn a small set of sampling points (\eg, four here) around the reference keypoint and further enlarge the feature spatial resolution while maintaining computational efficiency compared to vanilla DETR~\cite{carion2020detr}.
Each block is composed of a multi-head self-attention layer, a multi-scale deformable cross-attention layer, and FFNs. In the deformable cross-attention layer, keypoint queries $\mathbf{Q}$ extract features from the elements of multi-scale features $\mathbf{V}$ around the position of keypoints $p_q$:
\begin{equation}
    \text{CA}(\mathbf{Q}, \mathbf{V}, p_q)=\sum_{l=1}^L\sum_{k=1}^KA_{lqk} W\mathbf{V}_l(\phi_l(p_q)+\Delta p_{lqk}),
\end{equation}
where $l$ and $k$ index the feature level and keys, $A$ and $W$  are attention weight and learnable parameter. $\phi(\cdot)$ and $\Delta p$ are position rescaling and offset. After that, the updated component tokens $\mathbf{T_c}'\in \mathbb{R}^{K \times C'}$ will be fed into hand or face regression head to output the final hand or face parameters ($\hat{\mP}_{lhand}, \hat{\mP}_{rhand}, \hat{\mP}_{face}$), respectively.

\noindent \textbf{Loss Function.}
\modelname is trained in an end-to-end manner by minimizing the following loss function:
\begin{equation}
    L = L_{smplx} + L_{kpt3D} + L_{kpt2D} +L_{bbox2D}.
    \label{eq:loss}
\end{equation}
The four items are calculated as the L1 distance between the ground truth values and the predicted ones. Specifically, $L_{smplx}$ provides the explicit supervision of the SMPL-X parameters. $L_{kpt3D}$, $L_{kpt2D}$, and $L_{bbox2D}$ are regression losses for 3D whole-body keypoints, projected 2D whole-body keypoints, and left/right hands and face 2D bounding boxes. More details are provided in the Appendix.

%% file: sec/data.tex
\section{UBody--An Upper Body Dataset}
\label{sec:ubody}

3D whole-body mesh recovery from videos is a basic computer vision task, where it can provide comprehensive motion, gesture, and expression information to understand how humans perceive and act.
However, existing datasets lack scenes of downstream tasks, such as sign language recognition, gesture generation, emotion recognition, and real-life scenarios recorded as VLOGs, making recent state-of-the-art methods hard to generalize well on these scenes.
Interestingly, these scenarios are more concerned with the representations of \emph{upper bodies}.
We take this insight and present a novel large-scale benchmark for the expressive \emph{upper body} mesh recovery as shown in Figure~\ref{fig:ubody}(f) to (t), named \dataname. Our annotation pipeline is in Figure~\ref{fig:ubody_annotation}.
\emph{Due to the page limit, we put the data collection, data annotation processes, and annotation visualization in Appendix.} 

% \vspace{-0.1cm}
\subsection{Quality Analysis}
\label{sec:ubody_quality}

Our annotation pipeline produces far better 3D pseudo-GT fits with a shorter running time than the previous optimization-based and learning-based methods~\cite{HanbyulJoo2022eft,Feng_2021_pixie,Pavlakos_2019smplx,Moon_2022NeuralAnnot,pavlakos2022multishot}.
Figure~\ref{fig:ubody_vis_2d}(a) compares our 2D annotation results with the two wildly used annotation methods (OpenPose~\cite{ZheCao2018OpenPoseRM} and MediaPipe~\cite{FanZhang2020MediaPipeHO}). The quality of our 2D annotations is much more accurate, especially in terms of hand details and the robustness of occlusion and blur.
Figure~\ref{fig:ubody_vis_2d}(b) compares the 3D annotation of ours with the SOTA NeuralAnnot method~\cite{Moon_2022NeuralAnnot} on COCO. The quality of our approach is also better for the naked eye in terms of the fit of the body shape and the whole-body poses.

\subsection{Data Characteristics}
\label{sec:ubody_feature}

\begin{figure}[t]
\begin{center}
\includegraphics[width=1\linewidth]{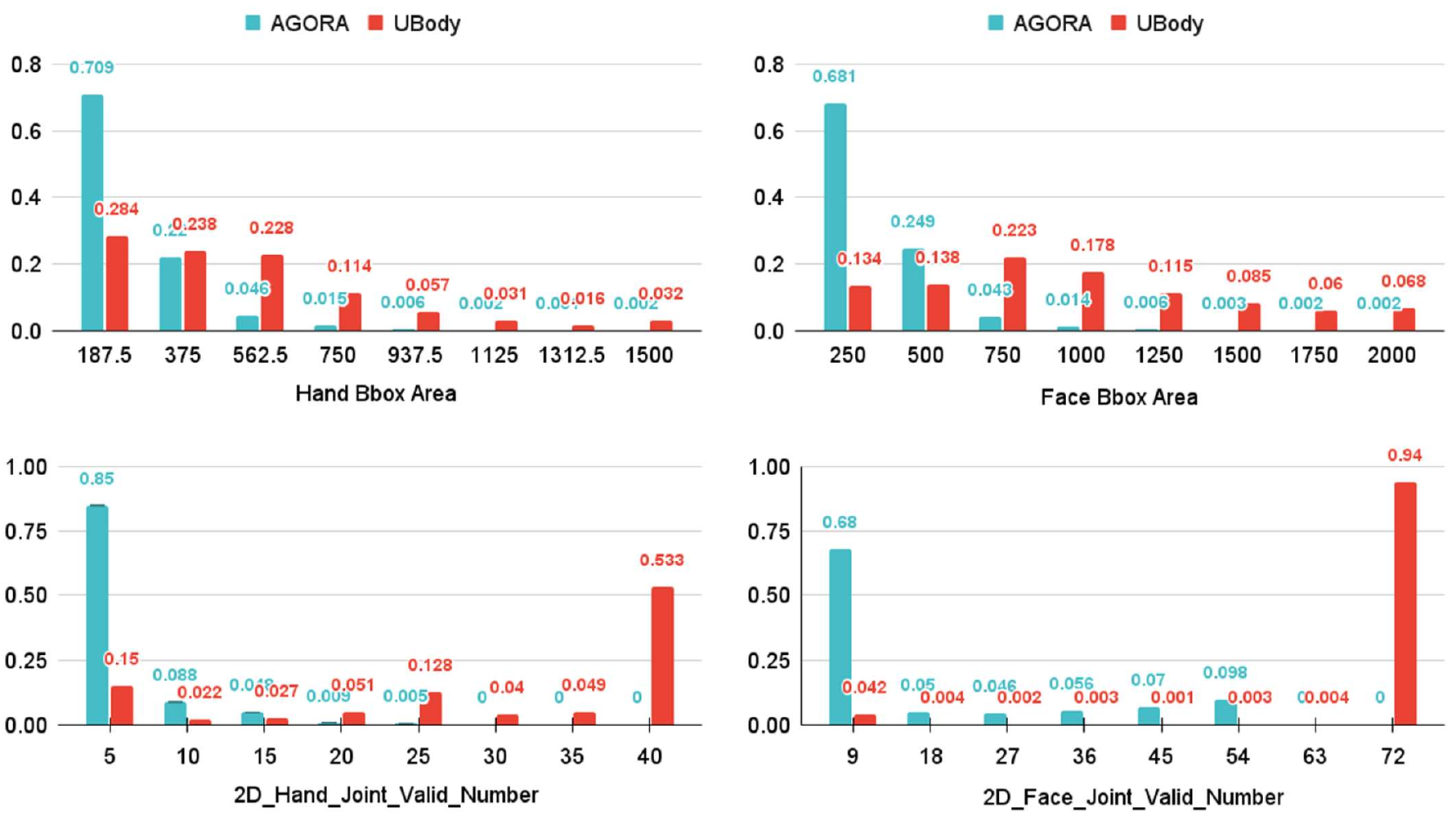}
\end{center}
\vspace{-0.6cm}
\caption{The statistical comparisons of the areas of the hand and face bounding boxes (upper row) and the number of 2D visible hand and face keypoints (lower row) with the logarithmic scale of the Y-axis. \dataname focuses on upper bodies exhibiting expressive gestures and facial expressions. 
}
\label{fig:statis}
\vspace{-0.6cm}
\end{figure}

Compared to the popular datasets illustrated in Figure~\ref{fig:ubody} (a) to (e) and the related human-centric datasets listed in Table~\ref{tab:datasets}, \dataname possesses unique features that present new challenges for future research.
%for existing methods.
%
Many videos are from edited media with highly diverse scenes and rich human actions and gestures. They have abrupt shot changes and dynamic camera viewpoints, leading to discontinuities between the frames. Close-up shots of humans cause severe truncation, making existing methods tend to fail. Meanwhile, they have varying degrees of interaction with objects and body components, subtitles, and special effects as occluded scenes. Also, there are high variations in background and light. Those conditions have not appeared in previous datasets.
All scenes in \dataname have rich hand gestures and facial expressions, making the recognition models pay more attention to these important body components. Lastly, all of these real-life videos provide audio as additional information to serve future multi-modality methods.
We also provide statistical comparisons between the key features of \dataname and the wildly used dataset AGORA~\cite{Patel_2021agora} in Figure~\ref{fig:statis}.
AGORA's hand/face bounding box area is generally small, while \dataname pays more attention to diverse hand and face scales as evidenced by its more dispersed area distribution. Meanwhile, \dataname has more visible face/hand keypoints, underscoring the importance of recognizing hand gestures and facial expressions. Lastly, \dataname's inclusion of  real-life videos provides new possibilities for subsequent spatio-temporal modeling that are not available in AGORA, which is an image-based dataset.

\vspace{-0.3cm}

%% file: sec/exp.tex
\vspace{0.1cm}
\section{Experiment}
\label{sec:exp}

\subsection{Experimental Setup}
Due to the page limit, we leave the detailed experiment setup, implementation, annotation visualization, qualitative comparison with SOTA methods, and more benchmark results and analyses in the appendix.

\noindent\textbf{Datasets.}
We use COCO-Wholebody~\cite{jin2020wholebody}, MPII~\cite{andriluka2014mpii}, and Human3.6M~\cite{Ionescu_2014_hm36} as the training set. Unlike previous multi-stage methods~\cite{PavlakosGeorgios2020expose, GyeongsikMoon2020hand4whole}, we do not use additional hand-only and face-only datasets for training as a simple baseline for a one-stage method. The SMPL/SMPL-X pseudo-GTs are obtained from EFT~\cite{joo2020exemplar} and NeuralAnnot~\cite{Moon_2022NeuralAnnot}. 

\noindent\textbf{Evaluation metrics.}
For 3D whole-body mesh recovery, we utilize the mean per-vertex position error (MPVPE) as our primary metric. In addition, we apply \emph{Procrustes Analysis} (PA) to the recovered mesh, and report the PA-MPVPE after rigid alignment. For AGORA, we also report normalized mean vertex error (N-PMVPE) to compensate for missing detection. Hand error is calculated as the mean of the left and right hands.
For 3D body-only recovery on 3DPW, we follow previous works~\cite{zeng2022deciwatch,Kolotouros_2019_spin} to report the mean per joint position error (MPJPE) and PA-MPJPE. 
All reported errors are in units of millimeters.

\noindent \textbf{Implementation details.}
\modelname is implemented in Pytorch and trained using the Adam optimizer with an initial learning rate of $1\times 10^{-4}$ for 14 epochs. Scaling, rotation, random horizontal flip, and color jittering are used as data augmentations during training. We set the number of body tokens $\mathbf{T}_b$ and component tokens $\mathbf{T}_c$ to 27 and 92, respectively.

\begin{table*}[h]
\centering
\resizebox{\textwidth}{!}
{
\begin{tabular}{l|ccc|cc|ccc|ccc|cc}
\specialrule{.1em}{.05em}{.05em}
\multirow{3}[4]{*}{\textbf{Method}} & \multicolumn{5}{c|}{\textbf{AGORA-test}} & \multicolumn{6}{c|}{\textbf{EHF}} &\multicolumn{2}{c}{\textbf{3DPW}} \\
\cmidrule{2-14}
& \multicolumn{3}{c|}{\textbf{MPVPE $\downarrow$}} & \multicolumn{2}{c|}{\textbf{N-MPVPE $\downarrow$}}& \multicolumn{3}{c|}{\textbf{MPVPE $\downarrow$}} & \multicolumn{3}{c|}{\textbf{PA-MPVPE $\downarrow$}}&\textbf{MPJPE $\downarrow$} & \textbf{PA-MPJPE $\downarrow$} \\
\cmidrule{2-14}
 & \textbf{All} & \textbf{~Hands~} & \textbf{~Face~} & \textbf{All} & \textbf{Body} & \textbf{All} & \textbf{Hands} & \textbf{Face} & \textbf{All} & \textbf{Hands} & \textbf{Face}& \textbf{Body}& \textbf{Body}\\ 
 \hline
ExPose~\cite{PavlakosGeorgios2020expose} &  217.3 & 73.1 & 51.1 & 265.0 & 184.8 & 77.1 & 51.6 & 35.0 & 54.5 & 12.8 & 5.8 & 93.4 & 60.7 \\
FrankMocap~\cite{Rong_2021frank} &  - & 55.2 & - & - & 207.8  &  107.6 & \underline{42.8} & - &57.5 & 12.6 & - & 96.7 & 61.9 \\
PIXIE~\cite{Feng_2021_pixie} &  191.8 & 49.3 & 50.2 & 233.9 & 173.4 & 89.2 & \underline{42.8} & 32.7 & 55.0 & \underline{11.1} & \textbf{4.6} & 91.0 & 61.3 \\

Hand4Whole~\cite{GyeongsikMoon2020hand4whole}  & - &-& -& -&- &79.2&	43.2	&\textbf{25.0} &53.1&	12.1&	5.8 &-&-  \\
Hand4Whole~\cite{GyeongsikMoon2020hand4whole}$\times$ & 135.5 & 47.2 & 41.6& 144.1&	96.0  & \underline{76.8} & \textbf{39.8} & \underline{26.1}&\textbf{50.3} & \textbf{10.8} & 5.8 &  \underline{86.6} & \underline{54.4}  \\
 \midrule
\modelname (Ours) &  \textbf{122.8}{\color{Red}$\downarrow_{9.5\%}$}& \textbf{45.7} & \textbf{36.2}&~~~~\textbf{130.6}~~~~&~~~~\textbf{85.3}~~~~&\textbf{70.8}{\color{Red}$\downarrow_{7.8\%}$} &53.7 &26.4 &\underline{48.7}&15.9&6.0&\textbf{74.7}{\color{Red}$\downarrow_{13.4\%}$} & \textbf{45.1} \\
 \specialrule{.1em}{.05em}{.05em}
\end{tabular}}
\vspace*{-3mm}
\caption{3D body reconstruction error comparisons on three existing datasets. $\times$ uses additional hand-only and face-only training datasets. }
\label{table:sota_compare}
\vspace{-0.2cm}
\end{table*}

\begin{table*}[h]
\begin{center}
\begin{minipage}[t]{0.405\textwidth}
\centering
\resizebox{0.95\linewidth}{!}
{
\begin{tabular}{l|cccc}
\toprule
\textbf{Hand}& ~\textbf{Ours}          & \textbf{w/o \emph{H.D.} }      & \textbf{w/o \emph{K.G}} & \textbf{w/o both }     \\
\midrule
\textbf{MPVPE }      & \textbf{53.7} & 55.3 & 55.1 &56.4 \\
\textbf{PA-MPVPE}   & \textbf{15.9} & 17.7 & 17.6 & 18.1 \\
\toprule
\textbf{Face}       & \textbf{Ours   }       & \textbf{w/o \emph{F.D.} }      & \textbf{w/o \emph{K.G}} & \textbf{w/o both}      \\
\midrule
\textbf{MPVPE}     & \textbf{26.4} & 27.2 & 26.4 & 26.8 \\
\textbf{PA-MPVPE}   & 6.0  & 5.9  & \textbf{5.8}  & 6.0 \\
\toprule
\textbf{Upsampling} & \textbf{$\times$ 1}&\textbf{$\times$ 2}&\textbf{$\times$ 4}&\textbf{$\times$ 8}\\
\midrule
\textbf{MPVPE} &54.9 & 54.3&\textbf{53.7} &54.1\\
\bottomrule
\end{tabular}}
\vspace{-0.2cm}
\caption{Ablation study of component-aware decoder on EHF with \emph{H.D.}, \emph{F.D.}, \emph{K.G}, and upsampling strategies. \emph{H.D.}, \emph{F.D.}, and \emph{K.G} are abbreviations for Hand Decoder, Face Decoder and Keypoint-Guided scheme. 
}
\label{tab:Break_Down}
\vspace{-0.7cm}
\end{minipage}
\quad
% Complexity
\begin{minipage}[t]{0.57\textwidth}
\centering
  \resizebox{\linewidth}{!}
{
    \begin{tabular}{l|ccc|ccc|cc}
    \toprule
    \multicolumn{1}{l|}{\multirow{2}[4]{*}{\textbf{Method}}} & \multicolumn{3}{c|}{\boldmath{}\textbf{MPVPE $\downarrow$}\unboldmath{}} & \multicolumn{3}{c|}{\boldmath{}\textbf{PA-MPVPE $\downarrow$}\unboldmath{}}&\multicolumn{2}{c}{\boldmath{}\textbf{PA-MPJPE $\downarrow$}\unboldmath{}}\\
    \cmidrule{2-9}
    & \textbf{All} & \textbf{Hand} & \textbf{Face} & \textbf{All} & \textbf{Hand} & \textbf{Face} &\textbf{Body} & \textbf{Hand} \\
    \midrule
    % SMPLify-X~\cite{Pavlakos_2019smplx} & & & & & & & & \\
    ExPose~\cite{PavlakosGeorgios2020expose} &171.5 & 83.7 & 45.1 & 66.9 & 12.0 & 3.9 &70.7	&12.3 \\
    % FrankMocap~\cite{Rong_2021frank} & & & & & & & & \\
    PIXIE~\cite{Feng_2021_pixie} &168.4 & 55.6 & 45.2 & 61.7 & 12.2  & 4.2 & 66.8	&12.3 \\
    Hand4Whole~\cite{GyeongsikMoon2020hand4whole} &104.1 & \underline{45.7} & 27.0 & 44.8 & \underline{8.9}  & 2.8 & 45.5&	\underline{9.0} \\
    %Hand4Whole~\cite{GyeongsikMoon2020hand4whole}$\dagger$ &	80.3&	40.9&	21.9&43.1&	7.4&	2.0&49.6&	- \\
    Hand4Whole~\cite{GyeongsikMoon2020hand4whole}$\times$ &157.4 & 62.2 & 49.8 & 82.2 & 9.8  & 3.9 & 92.8&	10.0\\
    \midrule
    \modelname (Ours) & \underline{92.4} & 47.7  &\underline{24.9} & \underline{42.4}& 10.8 & \underline{2.4}  & \underline{42.9}&	11.0 \\
    \modelname (Ours)$\dagger$ & \textbf{81.9}&	\textbf{41.5}&	\textbf{21.2}&\textbf{42.2}&\textbf{8.6}&	\textbf{2.0}&	\textbf{48.4}&	\textbf{8.8} \\
    \bottomrule
    \end{tabular}%
    }
    \vspace{-0.2cm}
  \caption[Reconstruction errors on the proposed \dataname test set.]{Reconstruction errors on \dataname test set on the \emph{intra-scene} protocol. All models are pretrained on previous datasets, except for the results labeled by (i) $\dagger$: finetuned on the \dataname training data; (ii) $\times$: finetuned on the AGORA training data. The result of the \emph{inter-scene} setting is in the appendix.}
  \label{tab:3d_smplx_results}%
  \end{minipage}
  \vspace{-0.9cm}
  \end{center}
\end{table*}

\subsection{Comparisons with Existing Methods}

\begin{figure}[t]
\vspace{-0.2cm}
\begin{center}
\includegraphics[width=1\linewidth]{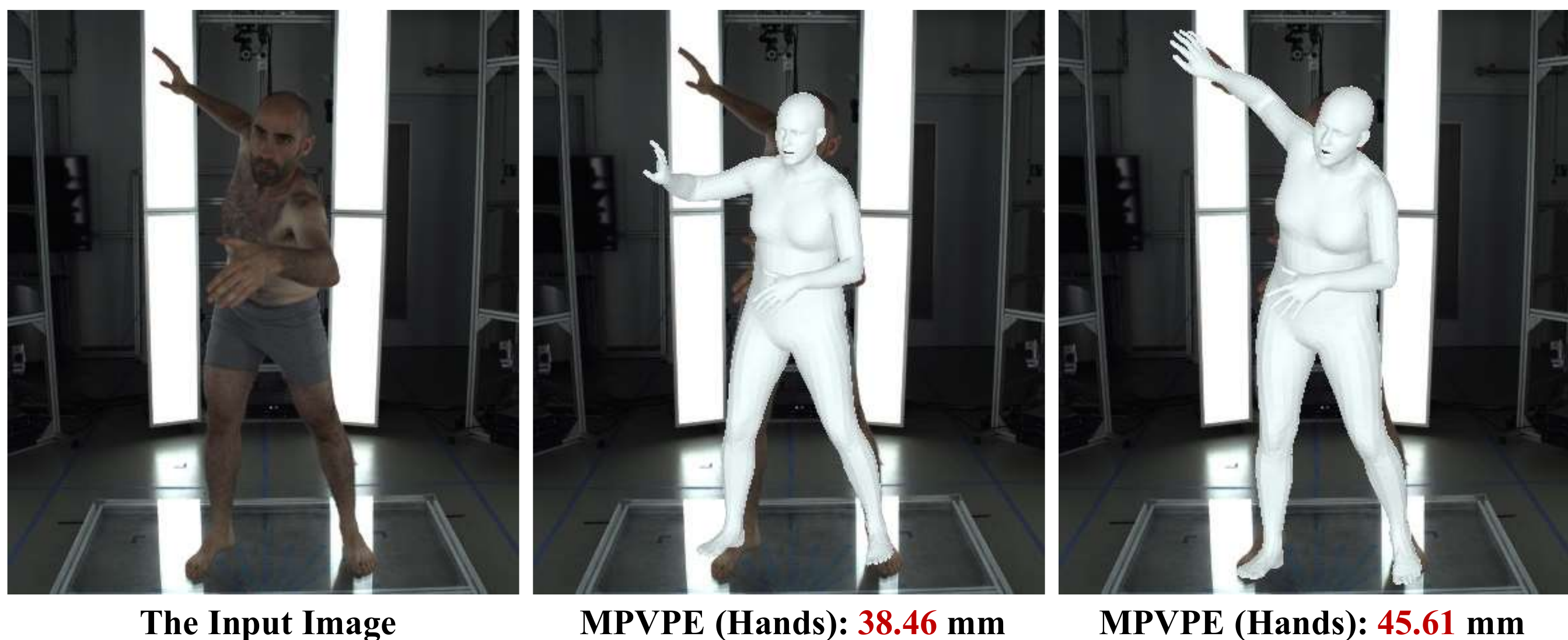}
\end{center}
\vspace{-0.4cm}
\caption{
Illustration of the inconsistency between quantitative and qualitative results compared Hand4Whole~\cite{GyeongsikMoon2020hand4whole} (the middle one) with \modelname (the right figure) on EHF.
}
\vspace{-0.6cm}
\label{fig:vis_ehf}
\end{figure}

Table~\ref{table:sota_compare} provides a comprehensive comparison of \modelname and existing whole-body mesh recovery methods. 
As the first one-stage method, \modelname surpasses existing multi-stage models with complex designs in most cases. Notably, \modelname has not been trained on hand-only and face-only datasets~\cite{Zimmermann_2019FreiHAND,TeroKarras2018ASG,GyeongsikMoon2020InterHand26MAD}. 
Our \emph{All MPVPEs} show a $9.5$\% improvement on AGORA test set and $7.8$\% improvement on EHF than SOTA~\cite{GyeongsikMoon2020hand4whole}.
Since AGORA is a more complex and natural dataset than EHF, previous works~\cite{GyeongsikMoon2020hand4whole,Patel_2021agora} claim it is more convincing and representative of real-world scenarios. We also visualize the misleading \emph{high-error} cases on EHF in Figure~\ref{fig:vis_ehf}.
Besides, we obtain a SOTA performance on the body-only dataset, 3DPW, with a $13.4$\% error reduction compared to these whole-body methods. More qualitative results are available in the appendix.

% \vspace{-0.1cm}
\subsection{Ablation Study}
\vspace{-0.1cm}
\noindent \textbf{Impact of the component-aware decoder.}
Unlike body-only pose estimation, whole-body mesh recovery requires attention to both the body's posture, which is on a larger spatial scale, and the gesture and expression of the hands and face, which are on a finer scale. To handle the resolution issue in a one-stage pipeline, we propose the component-aware decoder attached to the component-aware encoder.
First, in the upper Table~\ref{tab:Break_Down}, we verify the effectiveness of the proposed decoder for both hand and face regression. We can observe a significant drop without the decoder (\eg, \emph{w/o H.D.} and \emph{w/o F.D.}, indicating that simply regressing the low-resolution hand and face directly from the encoder is inferior. 
Moreover, the errors will also increase without the proposed keypoint-guided deformable attention scheme, as shown in the medium Table~\ref{tab:Break_Down}. In particular, the performance of the hand estimation is highly influenced, showing that hand pose estimation attends more to the sparsely deformable spatial information to obtain better queries.

\noindent \textbf{Impact of the up-sampling strategy.}
To relieve the low-resolution problem of hand and facial features, we design the feature up-sampling strategy in the decoder to obtain multi-scale higher-resolution features. 
The lower Table~\ref{tab:Break_Down} presents the impact of different up-sampling  scale. 
As the up-sampling scale increases, the MPVPE decreases and then reaches a saturation point. Therefore, we use three scales (i.e., $[\times 1, \times 2, \times 4]$) by default in our experiments.

\vspace{-0.2cm}
\subsection{Benchmark on UBody}
\vspace{-0.2cm}
As a new dataset, we provide both quantitative and qualitative results on \dataname. Table~\ref{tab:3d_smplx_results} presents the performance comparisons of existing 3D whole-body methods. The general result ranking is similar to AGORA. Since the upper body is closer to the camera, their errors will be smaller than AGORA. However, the hand and face will play a more important role than previous data. 
Besides, we finetune Hand4Whole on AGORA and test again, and we find all errors are significantly enlarged. This observation can be attributed to the data distribution gap between AGORA and \dataname, as shown in Figure~\ref{fig:statis}. Moreover, we train \modelname on our train set and find a 16.1\% improvement compared to the original pretrained model, indicating that \dataname can serve to improve the performance on downstream real-life scenes.

%% file: sec/conclusion.tex
\vspace{-0.1cm}
\section{Conclusion}
\vspace{-0.1cm}
In this work, we propose the first one-stage pipeline for 3D whole-body mesh recovery that achieves SOTA performance on three benchmarks in a simple yet effective manner. Moreover, to bridge the gap between the basic task of full-body pose and shape estimation and their downstream tasks, we develop a large-scale dataset with comprehensive scenes covering our daily life. With our proposed annotation method, we show that training on \emph{UBody} can effectively improve the performance of mesh recovery in upper-body scenes. We hope this work can contribute new insights to this area, both in terms of methodology and dataset.

\noindent \textbf{Limitation and future work.}
Currently, our training does not use additional hand and face-specific datasets. It is worth studying how to make the best use of them in our pipeline to further improve performance. Also, we can validate the effectiveness of \emph{UBody} on some downstream applications, \eg, gesture recognition, driving avatar.

\textbf{Acknowledgements:} This work was partially funded through the National Key Research and Development Program of China (Project No.2022YFB36066), in part by the Shenzhen Science and Technology Project under Grant (CJGJZD20200617102601004, JCYJ20220818101001004).

%% file: sec_sup/supp.tex
% \title{\emph{Supplementary Material}\\ One-Stage 3D Whole-Body Mesh Recovery with Component Aware Transformer}  % **** Enter the paper title here
% \maketitle
% \thispagestyle{empty}
% \appendix

% \null\vfill % Add vertical space at the top of the page
% \begin{center}
% \begin{minipage}{\textwidth} % Use a minipage to keep the text centered
%   \centering
%   \textbf{\emph{Supplementary Material}\\ One-Stage 3D Whole-Body Mesh Recovery with Component Aware Transformer}
% \end{minipage}
% \end{center}
% \vfill\null % Add vertical space at the bottom of the page
\setcounter{section}{0}
\setcounter{table}{0}
\setcounter{figure}{0}

\renewcommand{\thesection}{\Alph{section}}   
\renewcommand {\thetable} {S-\arabic{table}}
\renewcommand {\thefigure} {S-\arabic{figure}}

\pagebreak

\twocolumn[{
	\renewcommand\twocolumn[1][]{#1}
	\begin{center}
		\textbf{\Large Supplementary Material:\\One-Stage 3D Whole-Body Mesh Recovery with Component Aware Transformer}
        \vspace{0.8cm}
        \end{center}
}]

\section*{Overview}
\noindent This supplementary material presents more details and additional results not included in the main paper due to page limitation. The list of items included are:

\begin{itemize}
    \item More experiment setup and details in Sec.~\ref{sec:exp_setup}.
    \item Efficiency comparison with SOTA in Sec.~\ref{sec:efficiency_comp}.
    \item Experiment on AGORA dataset in Sec.~\ref{sec:agora}. 
    \item More introduction of UBody in Sec.~\ref{sec:ubody_intro}. 
    \item Inter-scene benchmark on UBody dataset in Sec.~\ref{sec:inter_bench}. 
    \item Qualitative comparisons with SOTA in Sec.~\ref{sec:visual}.
\end{itemize}

\blfootnote{$\S$ Work done during an internship at IDEA; ${\P}$~Corresponding author.}\

\section{Experiment Setup}
\label{sec:exp_setup}

\noindent\textbf{Evaluation metrics.} To quantitatively evaluate the performance of human mesh recovery, MPVPE, PA-MPVPE, MPJPE, and PA-MPJPE are used as evaluation metrics. Besides, we also report normalized mean vertex error (NMVE) and normalized mean joint error (NMJE) by the standard detection metric, F1 score (the harmonic mean of recall and precision) to penalize models for misses and false positives on AGORA test set with many multi-person scenes.

\noindent\textbf{Implementation details.} Our \modelname model is implemented in Pytorch. It is trained with Adam optimizer ($\beta_1=0.1, \beta_2=0.999$) using the Cosine Annealing scheme for 14 epochs. The learning rate is initially set to $1\times10^{-4}$. The batch size is set to 192. Random scaling, rotation, horizontal flip, and color jittering are used as data augmentations during training. The spatial size of the input image is $256\times 192$. The number of body tokens $\mathbf{T}_b$ and component tokens $\mathbf{T}_c$ are set to 27 and 92, respectively. During experiments on the AGORA-test set, we remove the decoder as we find that the decoder increases training time and does not significantly improve performance on AGORA-test set. This observation may be attributed to the fact that the main problem of AGORA is occlusion, while the decoder aims to estimate hands/face at a finer level.

\section{Efficiency comparison with SOTA methods}
\label{sec:efficiency_comp}
\noindent We report the complexity comparisons including average inference time, number of model parameters, FLOPs, and the NMJE-All on AGORA-test in Table~\ref{tab:efficiency_comparison}. The numbers are measured for single-person regression on the same resolution input using a machine with an NVIDIA A100 GPU. OSX has \emph{the shortest inference time and lowest error}, indicating the advantages in practical applications.
        
\begin{table}[h]
    \begin{center}
        \vspace{-3.0mm}
        \hspace{-1mm}
        \scalebox{0.67}{\noindent
            \begin{tabular}{c | c c c c c }
                \toprule
                Method &ExPose~[\textcolor{green}{34}]  & PIXIE~[\textcolor{green}{13}] & H4W~[\textcolor{green}{27}] & PyMAF-X~[\textcolor{green}{48}] & OSX \\
                \midrule
                NMJE-All (mm) &263.3  &230.9  &141.1  &140.0 & 127.6    \\
                Infer Time (ms) &120.2 &192.0 & 73.3 &209.3 & 54.6 \\
                ~Params (M)~ &135.8 &192.9 &77.9 &205.9 & 102.9 \\
                ~FLOPS (G)~ &28.5 &34.3 &16.7 &35.5 & 25.3 \\
                %\midrule
                \bottomrule
        \end{tabular}}
        \vspace{-3.0mm}
        \caption{Efficiency comparisons with multi-stage methods. }
        \label{tab:efficiency_comparison}
    \end{center}\vspace{-8mm}
\end{table}

\begin{table*}[h]
  \centering
  \resizebox{\textwidth}{!}
{
    \begin{tabular}{l|cc|cc|cccc|cccc}
    \toprule
    \multicolumn{1}{c|}{\multirow{2}[4]{*}{\textbf{Method}}}  & \multicolumn{2}{c|}{\boldmath{}\textbf{NMVE $\downarrow$}} & 
    \multicolumn{2}{c|}{\boldmath{}\textbf{NMJE $\downarrow$}} & 
    \multicolumn{4}{c|}{\boldmath{}\textbf{MVE $\downarrow$}\unboldmath{}} & \multicolumn{4}{c}{\boldmath{}\textbf{MPJPE $\downarrow$}\unboldmath{}} \\
    \cmidrule{2-13} & \textbf{Full-Body} & \textbf{Body} & \textbf{Full-Body} & \textbf{Body} &
    \textbf{Full-Body} & \textbf{Body} & \textbf{Face} & \textbf{LH/RH} & \textbf{Full-Body} & \textbf{Body} & \textbf{Face} & \textbf{LH/RH} \\
    \midrule
    SMPLify-X~\cite{Pavlakos_2019smplx} & 333.1  & 263.3 & 326.5 & 256.5 & 236.5 &187.0 & 48.9  & 48.3/51.4 & 231.8 & 182.1 & 52.9  & 46.5/49.6 \\
    ExPose~\cite{PavlakosGeorgios2020expose} & 265.0 & 184.8 &263.3 & 183.4  & 217.3 & 151.5 & 51.1  & 74.9/71.3 & 215.9 & 150.4 & 55.2  & 72.5/68.8 \\
    FrankMocap~\cite{Rong_2021frank} & - & 207.8 & - & 204.0  & - & 168.3 & -     & 54.7/55.7 & -     & 165.2 & -     & 52.3/53.1 \\
    PIXIE~\cite{Feng_2021_pixie} &  233.9 & 173.4 & 230.9 & 171.1  & 191.8 & 142.2 & 50.2  & 49.5/49.0 & 189.3 & 140.3 & 54.5  & 46.4/46.0 \\
    Hand4Whole~\cite{GyeongsikMoon2020hand4whole} $^\dagger$ & 144.1 & 96.0 & 141.1 & \underline{92.7} & 135.5 & 90.2  & 41.6  & 46.3/48.1 & 132.6 & 87.1  & 46.1  & 44.3/46.2 \\
    PyMAF-X~\cite{HongwenZhang2022PyMAFXTW} $^\dagger$ & \underline{141.2} & \underline{94.4} & \underline{140.0} & 93.5 & \underline{125.7} & \underline{84.0} & \textbf{35.0} & \textbf{44.6/45.6} & \underline{124.6} & \underline{83.2} & \textbf{37.9} & \textbf{42.5/43.7} \\
    \midrule
    OSX (Ours) $^\dagger$ & \textbf{130.6}{\color{Red}$\downarrow_{7.5\%}$} & \textbf{85.3}{\color{Red}$\downarrow_{9.6\%}$} & \textbf{127.6}{\color{Red}$\downarrow_{8.9\%}$} & \textbf{83.3}{\color{Red}$\downarrow_{10.9\%}$} & \textbf{122.8} & \textbf{80.2} & \underline{36.2} & \underline{45.4/46.1} & \textbf{119.9} & \textbf{78.3} & \textbf{37.9} & \underline{43.0/43.9}  \\
    \bottomrule
    \end{tabular}%
}
  \caption[Reconstruction errors on the AGORA val set.]{Reconstruction errors on the AGORA test set. $^\dagger$ denotes the methods that are fine-tuned on the AGORA training set or similarly synthetic data~\cite{kocabas2021spec}. The best results are shown in \textbf{bold} and the second best results are highlighted with \underline{underlined font}.}
  \label{tab:agora_test}%
\end{table*}%

\section{Experiment on AGORA Dataset}
\label{sec:agora}
In this part, we report the complete result on the AGORA test set and the experiment result on the AGORA val set.

\noindent\textbf{AGORA Test Set.} Table~\ref{tab:agora_test} depicts the complete result on the AGORA test set. All the results are taken from the official leaderboard. As shown, our OSX outperforms other competitors on most metrics, especially on the evaluation of the body and full-body recovery. More specifically, for full-body reconstruction, OSX even surpasses PyMAF-X~\cite{HongwenZhang2022PyMAFXTW} by 10.6 mm, 9.1 mm, 2.9 mm, and 4.7 mm on NMVE, NMJE, MVE, and MPJPE, respectively. Since PyMAF-X has a lower detected person ratio, they have similar results on MVE and MPJPE metrics, which only calculate the matched person. The NMVE and NMJE will take the misses and false positives into account, and we have overall better multi-person estimation with more improvement under the metrics.
Notably, although OSX does not use extra hand-only and face-only datasets, it can achieve competitive results on hand and face metrics, which demonstrates the effectiveness of our component-aware decoder.

\noindent\textbf{AGORA Val Set.} Table~\ref{tab:agora_val} shows the result on the AGORA val set. All the results are taken from ~\cite{GyeongsikMoon2020hand4whole} except OSX. Although we do not use extra hand/face specific datasets during training, OSX outperforms the SOAT method Hand4Whole by 8.3\% on the MPVPE-all, demonstrating the effectiveness of our one-stage method. 

\begin{table}[h]
\centering
  \resizebox{\linewidth}{!}
{
    \begin{tabular}{l|ccc|ccc}
    \toprule
    \multicolumn{1}{c|}{\multirow{2}[4]{*}{\textbf{Method}}} & \multicolumn{3}{c|}{\boldmath{}\textbf{MPVPE $\downarrow$}\unboldmath{}} & \multicolumn{3}{c}{\boldmath{}\textbf{PA-MPVPE $\downarrow$}\unboldmath{}}\\
    \cmidrule{2-7}
    & \textbf{All} & \textbf{Hand} & \textbf{Face} & \textbf{All} & \textbf{Hand} & \textbf{Face} \\
    \midrule
    % SMPLify-X~\cite{Pavlakos_2019smplx} & & & & & & & & \\
    ExPose~\cite{PavlakosGeorgios2020expose} &219.8 & 115.4 & 103.5 & 88.0 & 12.1 & 4.8 \\
    FrankMocap~\cite{Rong_2021frank} & 218.0 & 95.2 & 105.4 & 90.6 &11.2 &4.9  \\
    PIXIE~\cite{Feng_2021_pixie} &203.0 & 89.9 & 95.4 & 82.7 & 12.8  & 5.4  \\
    Hand4Whole~\cite{GyeongsikMoon2020hand4whole} &183.9 &	72.8	&81.6 & 73.2 & \textbf{9.7} &	\textbf{4.7} \\
    \textbf{OSX (Ours)} &	\textbf{168.6\color{Red}$\downarrow_{8.3\%}$}&	\textbf{70.6}&	\textbf{77.2} & \textbf{69.4}&	\underline{11.5}& 4.8 \\
    \bottomrule
    \end{tabular}%
    }
    \vspace{-0.2cm}
  \caption[]{Reconstruction errors on the AGORA val set.}
  \label{tab:agora_val}%
  \vspace{-0.7cm}
\end{table}

\section{UBody: An Upper Body Dataset}
\label{sec:ubody_intro}

%示意每个场景我们的一张图的标注质量

\subsection{Data Collection}
To bridge the gap between the basic human mesh recovery task and its downstream applications, we design \dataname with two rules. 
First, we research a wide range of human-related downstream tasks with upper-body scenes, including gesture recognition~\cite{yoonICRA19,guo2021human,mitra2007gesture}, sign language recognition, and translation~\cite{duarte2021how2sign,rastgoo2021sign,subburaj2022survey,Joze2019MSASLAL,camgoz2021content4all,zhou2021spatial}, person clustering~\cite{brown2021face}, emotion analysis, speaker verification~\cite{nagrani2020voxceleb}, micro-gesture understanding~\cite{liu2021imigue}, audio-visual generation and separation~\cite{pu2017audio}, human action recognition, and localization~\cite{pavlakos2022multishot,ChunhuiGu2017AVAAV,rockwell2020full,siarohin2021motion,fouhey2018lifestyle}, and human video segmentation~\cite{kuang2021flow}. We select the corresponding high-quality datasets from these existing tasks as a part of our data for the corresponding scenarios. In order to ensure a balanced amount of data for each scene, for datasets with many videos (\eg, lasting 20k minutes), we manually selected the videos in which the upper body appeared more frequently.

Second, with all kinds of athletic competitions, entertainment shows, we media, online conferences, and online classes being more and more indispensable, we carefully selected a large number of rich videos from YouTube to provide new opportunities and challenges for potential applications. 

Since some untrimmed videos may have missing main characters, extraneous images such as opening and closing credits, and repetitive actions, we manually fine-cut the long videos. Each edited video is 10 seconds long, which ensures the high quality of the video.

In order to prevent infringement of ownership rights, we only provide download links to the corresponding videos and our labels without any personal information.

In summary, we collect fifteen real-life scenarios with more than \textbf{105,1k} frames. 
%\TODO{How many clips?}. 
We split the train/test sets from two protocols as follows. 
\begin{itemize}
\item
\emph{Intra-scene}: in each scene, the former 70\% of the videos are the training set, and the last 30\% are the test set. The benchmark was provided in the main paper.
\item
\emph{Inter-scene}: we use ten scenes of the videos as the training set and the other five scenes as the test set. Due to the page limit, we present the benchmark in Table~\ref{tab:3d_smplx_results_sup}.
%\TODO{What is the usage of this? We do not use this config in main manuscript, right?}
\end{itemize}

\subsection{Data Annotation Processes}
\label{sec:ubody_annotation}

\begin{figure*}[h]
\centering
\includegraphics[width=1\linewidth]{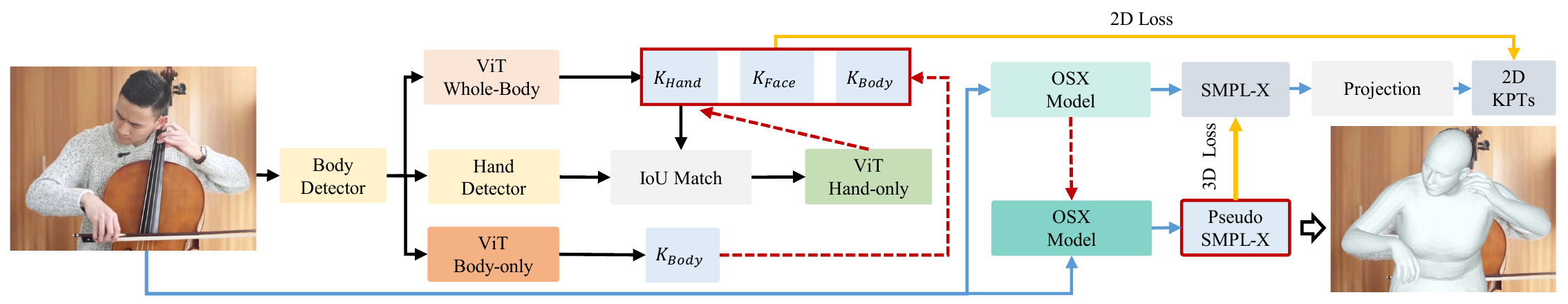}
\vspace{-0.5cm}
\caption{
Illustration of the annotation pipeline of \dataname. Black lines show the annotation process of 2D whole-body keypoints, and blue lines are the 3D SMPL-X annotation procedure. Red dotted lines mean to update the information.
}
\label{fig:ubody_annotation_sup}
\vspace{-0.3cm}
\end{figure*}

As shown in Figure~\ref{fig:ubody_annotation_sup}, we design a thorough whole-body annotation pipeline with high precision. It is divided into two stages: 2D whole-body keypoint annotation and 3D SMPLX annotations fitting.
Since \dataname scenes have a number of unpredictable transitions and cutscenes that make it difficult to use the temporal smoothing approaches~\cite{young1995gaus1d,press1990savitzky,zeng2022smoothnet}, the annotation is conducted on a single frame.

\noindent \textbf{2D whole-body keypoint annotation:} 
%
%The definition of $133$ 2D keypoint follows COCO-wholeBody~\cite{jin2020wholebody}. 
We first detect all persons and their hands in an image via a specific human and hand detector \emph{BodyHands}~\cite{narasimhaswamy2022bodyhands} shown as \emph{Body Detector} and \emph{Hand Detector} in Figure~\ref{fig:ubody_annotation}.
Leveraging the recent state-of-the-art 2D pose estimator \emph{ViT-Body-only}~\cite{YufeiXu2022ViTPoseSV}, we use the pre-trained model trained on the COCO~\cite{lin2014coco} dataset to localize 17 body keypoints for each detected single person, named \emph{$K_{Body}$}, which shows highly robust results on many scenes.
Due to the diverse scales and motion blur for the fast-moving hands, we find that \emph{Hand Detector} will output false positive samples or miss some hands. To enhance the performance of hand detection, 
we train a 2D whole-body estimator on COCO-wholeBody~\cite{jin2020wholebody} with $133$ 2D keypoints, called \emph{ViT-WholeBody} following the model design of ViTPose~\cite{YufeiXu2022ViTPoseSV} and masked autoencoder pre-trained scheme~\cite{he2022masked}. \emph{ViT-WholeBody} can provide high-recall hand keypoints \emph{$K_{Hand}$}, but the localization precision is low because of the fully one-stage pipeline and low-resolution of hands from the raw image. Accordingly, We can obtain coarse hand bounding boxes by calculating the maximum, and minimum values of the detected left and right-hand keypoints to correct the hand boxes from \emph{Hand Detector} via an IoU matching strategy.
Then, we use the fine hand boxes to crop the hand patches, resize them to a larger size, and put them into our specific pre-trained \emph{ViT-Hand-only} model trained with the hand labels from the COCO-Whole dataset.
In summary, \emph{ViT-WholeBody} will output the body, hand, and face 2D keypoints. We use the body output from \emph{ViT-Body-only} to replace the \emph{$K_{Body}$}, and use the fine hand keypoints from \emph{ViT-Hand-only} to change the \emph{$K_{Hand}$}.
As the face of the current SMPL-X model does not require much detail, we simply use the 2D face keypoints \emph{$K_{Face}$} obtained from \emph{ViT-WholeBody}.

\noindent\textbf{3D whole-body mesh recovery annotation:} 
Different from previous optimization-based annotation~\cite{Pavlakos_2019smplx} that may output implausible poses, we use our proposed \modelname to estimate the SMPL-X parameters from human images as a proper 3D initialization to provide pseudo-3D constraints.
Benefiting from current 2D keypoint localization that tends to be more accurate, we additionally supervise the projected 2D whole-body keypoints by the above annotated 2D whole-body keypoints as a way to train \modelname.
More importantly, to avoid performance degradation from not accurate enough initial labeling and consistently push up the 3D annotation quality, we propose an iterative training-labeling-revision loop for every 30 epochs to train 120 epochs in total. 
% Next, we project the 3D information into a 2D coordinate and obtain the corresponding 2D whole-body keypoints. We can supervise 

% \TODO{Add the difference between previous works}

\section{Inter-Scene Benchmark on UBody dataset}
\label{sec:inter_bench}

Due to the page limit, we further provide another data protocol comparison to show the usage of the proposed \dataname. Table~\ref{tab:3d_smplx_results_sup} presents the performance comparisons of existing 3D whole-body methods. Inter-scene test shows large errors than the intra-scene test due to the different motion and gesture distributions.
The model finetuned on AGORA still has a significant gap than trained on the COCO dataset.
Furthermore, we also train Hand4Whole and \dataname on our training set, we can find a consistent improvement compared to the original pretrained model, indicating that \dataname can serve to bridge the gap among these downstream real-life scenes. 
Moreover, different from single-frame AGORA and EHF, \dataname provides videos, which can drive progress in spatial-temporal modeling on such edit media sources.

\begin{table}[h]
\centering
  \resizebox{\linewidth}{!}
{
    \begin{tabular}{l|ccc|ccc}
    \toprule
    \multicolumn{1}{c|}{\multirow{2}[4]{*}{\textbf{Method}}} & \multicolumn{3}{c|}{\boldmath{}\textbf{MPVPE $\downarrow$}\unboldmath{}} & \multicolumn{3}{c}{\boldmath{}\textbf{PA-MPVPE $\downarrow$}\unboldmath{}}\\
    \cmidrule{2-7}
    & \textbf{All} & \textbf{Hand} & \textbf{Face} & \textbf{All} & \textbf{Hand} & \textbf{Face} \\
    \midrule
    % SMPLify-X~\cite{Pavlakos_2019smplx} & & & & & & & & \\
    ExPose~\cite{PavlakosGeorgios2020expose} &185.7 & 89.5 & 47.2 & 76.4 & 11.8 & 4.0 \\
    % FrankMocap~\cite{Rong_2021frank} & & & & & & & & \\
    PIXIE~\cite{Feng_2021_pixie} &185.0 & 60.9 & 45.3 & 74.5 & 11.9  & 4.2  \\
    Hand4Whole~\cite{GyeongsikMoon2020hand4whole}$\times$ &198.1&	66.9	&51.8 & 90.2&	10.3&	4.1 \\
    Hand4Whole~\cite{GyeongsikMoon2020hand4whole} &109.4 & {50.4} & 24.8 & 57.0 & {8.9}  & 2.7 \\
    Hand4Whole~\cite{GyeongsikMoon2020hand4whole}$\dagger$ &	\underline{87.4}&	\textbf{41.6}&	\underline{22.1}&\underline{46.3}&	\textbf{8.0}&\underline{2.0} \\
    
    \midrule
    \modelname (Ours) & {100.7} & 52.5  &{24.5} & {52.9}& 9.5 & {2.6}   \\
    \modelname (Ours)$\dagger$ & \textbf{82.0}&	\underline{44.2}&	\textbf{21.5}&\textbf{44.2}&\underline{8.8}&	\textbf{1.9} \\
    \bottomrule
    \end{tabular}%
    }
    \vspace{-0.2cm}
  \caption[Reconstruction errors on the proposed \dataname test set.]{Reconstruction errors on \dataname test set on the \emph{inter-scene} protocol. All models are pretrained on previous datasets, except for the results labeled by (i) $\dagger$: finetuned on the \dataname training data; (ii) $\times$: finetuned on the AGORA training data. }
  \label{tab:3d_smplx_results_sup}%
  % \vspace{-0.2cm}
\end{table}

\section{Qualitative with SOTA method}
\label{sec:visual}

\begin{figure*}[h]
\begin{center}
\includegraphics[width=1\linewidth]{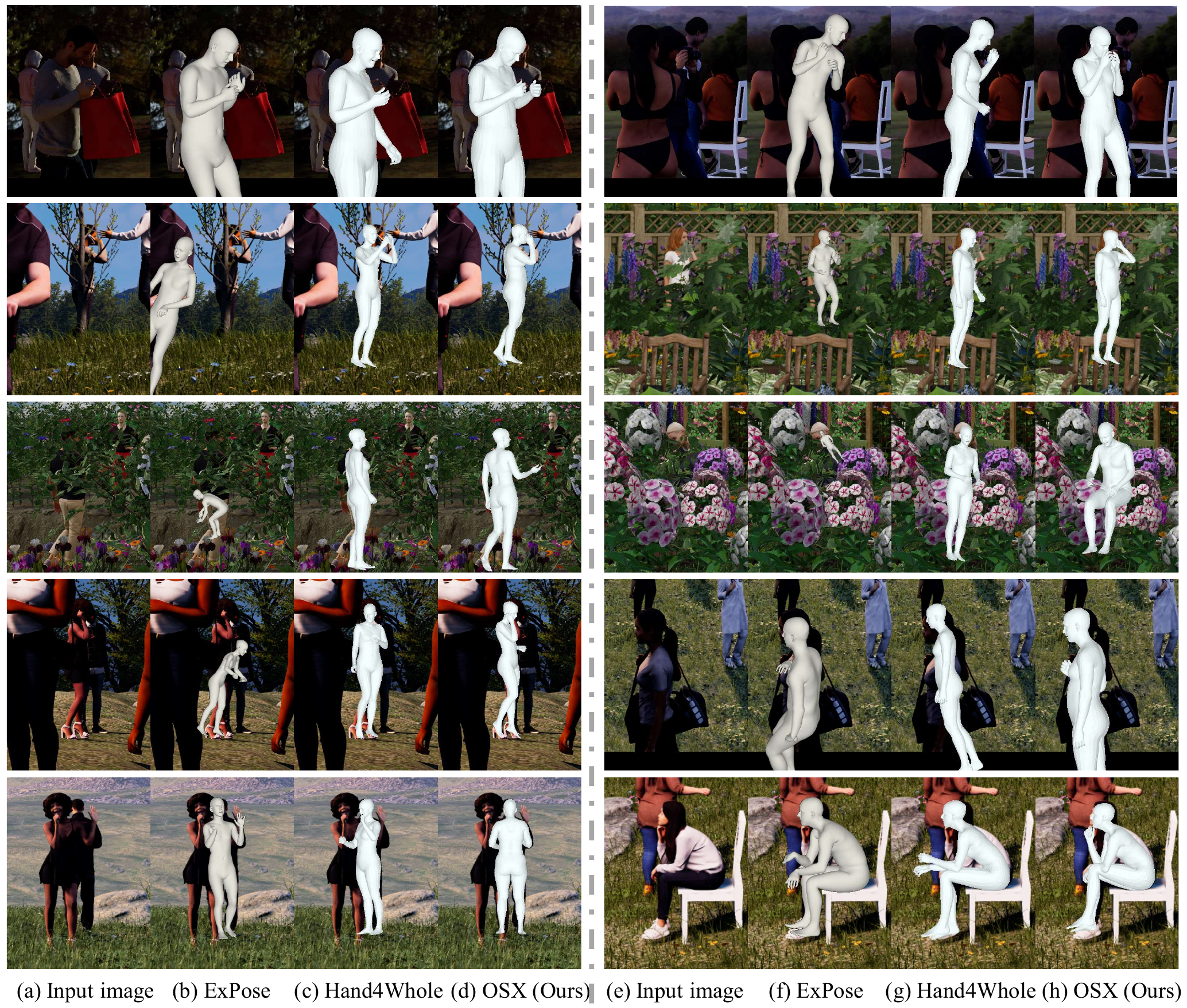}
\end{center}
\vspace{-0.6cm}
\caption{
Comparisons of existing 3D whole-body estimation methods on AGORA.
}
\label{fig:vis_agora}
\end{figure*}

\noindent\textbf{Qualitative comparisons on AGORA:} 
We compare the mesh quality on the AGORA dataset in Figure~\ref{fig:vis_agora}. Agora is a synthetic dataset with many challenging factors like heavy occlusion, dark environment, and unnatural multi-person interaction. It only has limited actions, \eg, taking phones, walking, sitting,  \etc. We can see \modelname outperforms ExPose~\cite{PavlakosGeorgios2020expose} and Hand4Whole~\cite{GyeongsikMoon2020hand4whole} consistently in terms of global body orientations, whole-body poses, and hand pose.

\begin{figure*}[h]
\begin{center}
\includegraphics[width=0.9\linewidth]{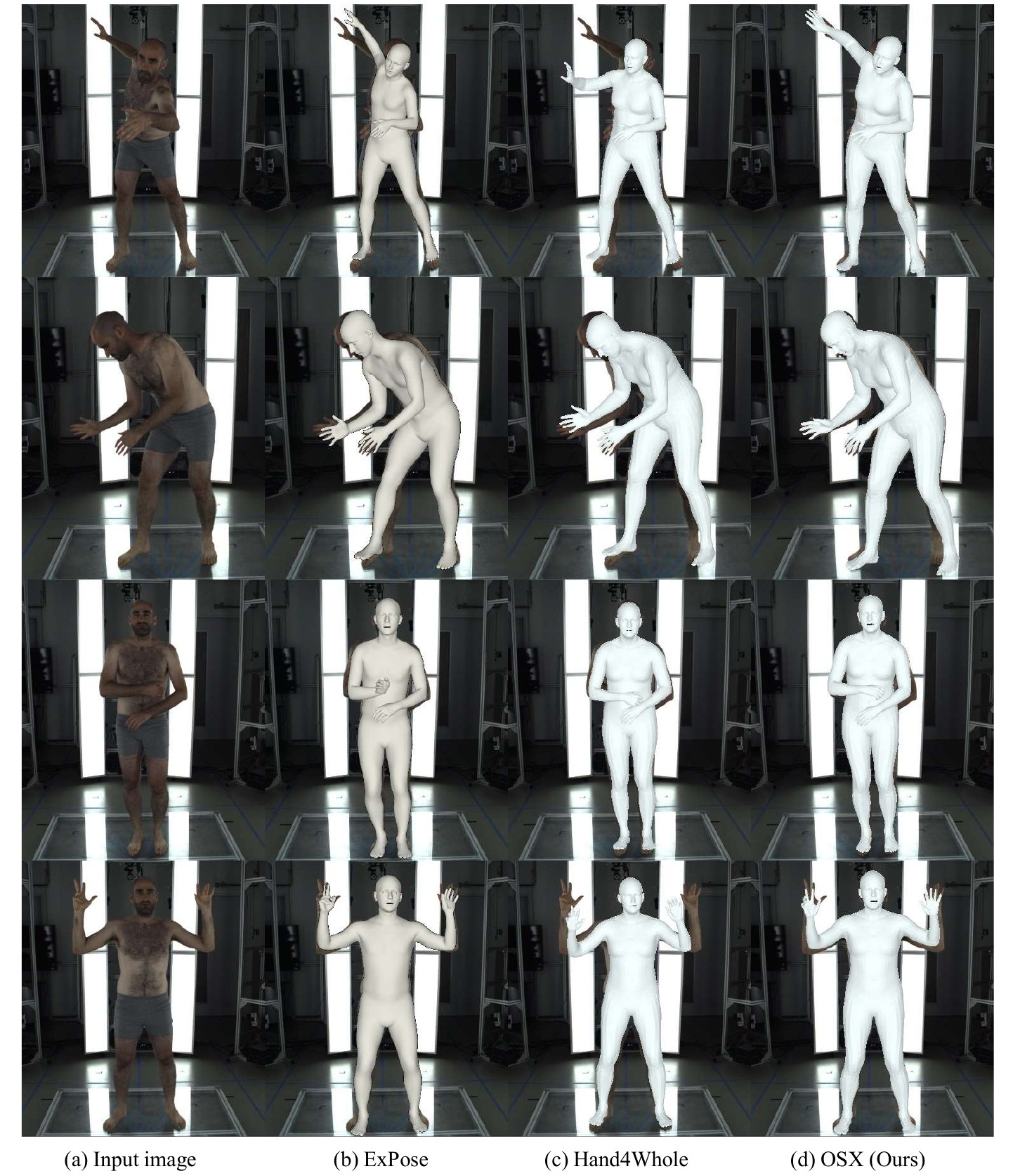}
\end{center}
\vspace{-0.6cm}
\caption{
Comparisons of existing 3D whole-body estimation methods on EHF.
}
\label{fig:vis_ehf_sup}
\end{figure*}

\noindent\textbf{Qualitative comparisons on EHF:} 
The visual comparisons of whole-body mesh recovery quality on the EHF dataset can be found in Figure~\ref{fig:vis_ehf_sup}. As can be seen, \modelname estimates the most accurate whole-body poses, in which the body parts like hands, feet, and hands are better aligned with the person in the image. 

\begin{figure*}[h]
\begin{center}
\includegraphics[width=1\linewidth]{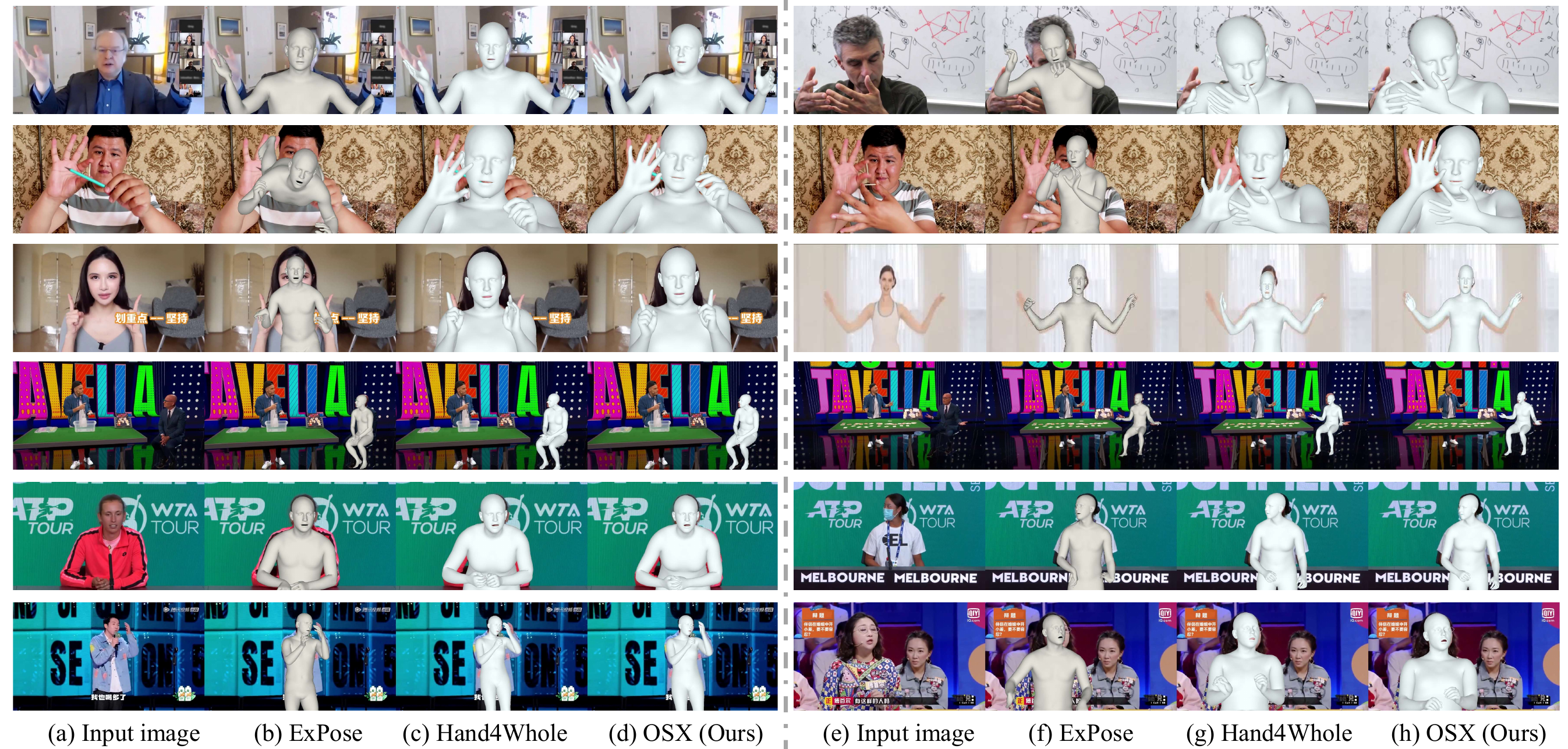}
\end{center}
\vspace{-0.6cm}
\caption{
Comparisons of existing 3D whole-body estimation methods on our proposed \dataname.
}
\label{fig:vis_ubody}
\end{figure*}

\noindent\textbf{Qualitative comparisons on UBody}: 
The qualitative comparison on our \dataname is in Figure~\ref{fig:vis_ubody}. \dataname focuses more on the expressive upper body part. Hand4Whole~\cite{GyeongsikMoon2020hand4whole} and our \modelname produces better body mesh recoveries than ExPose~\cite{YufeiXu2022ViTPoseSV}. Close inspection of the hand part shows that our hand recovery is more accurate than Hand4Whole. 

\begin{figure*}[h]
\begin{center}
\includegraphics[width=1\linewidth]{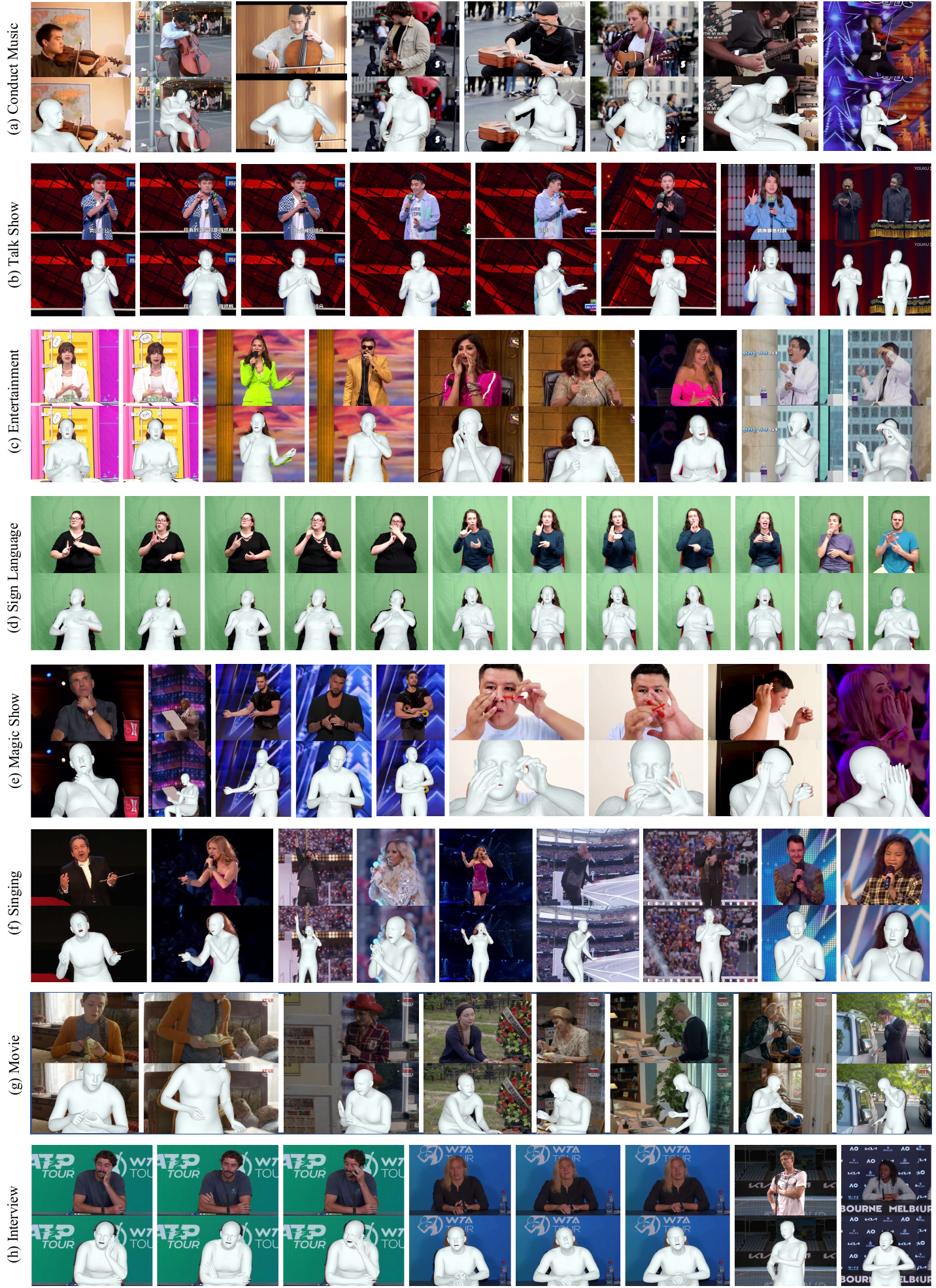}
\end{center}
\vspace{-0.6cm}
\caption{
Illustration of the ground-truth SMPL-X annotation for the eight scenes in \dataname. For each scene, we show the input image (the upper) and our annotation (the lower).
}
\label{fig:vis_ubody_gt1}
\end{figure*}

\begin{figure*}[h]
\begin{center}
\includegraphics[width=1\linewidth]{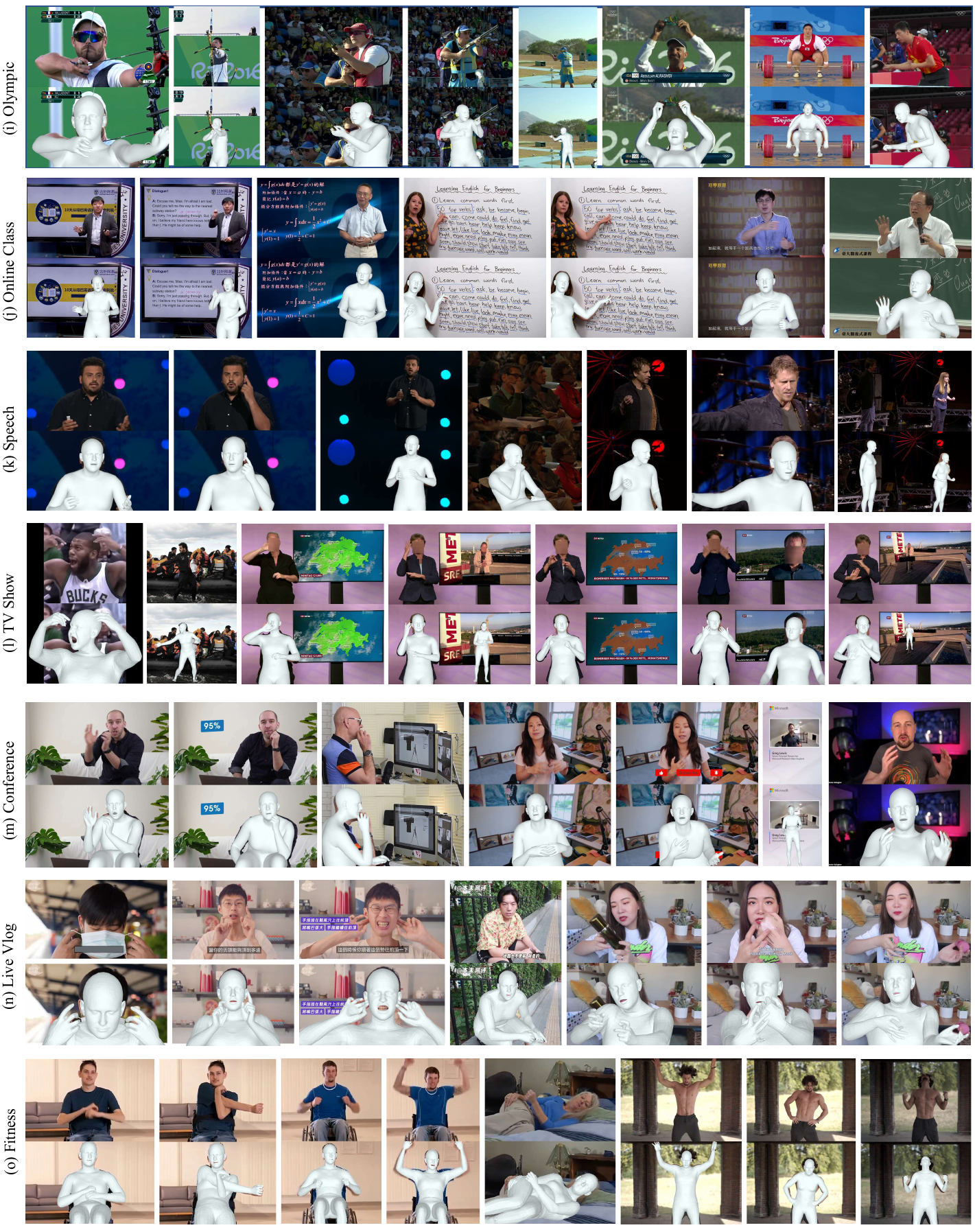}
\end{center}
\vspace{-0.6cm}
\caption{
Illustration of the ground-truth SMPL-X annotation for seven other scenes in \dataname. For each scene, we show the input image (the upper) and our annotation (the lower).
}
\label{fig:vis_ubody_gt2}
\end{figure*}

\begin{figure*}[h]
\begin{center}
\includegraphics[width=1\linewidth]{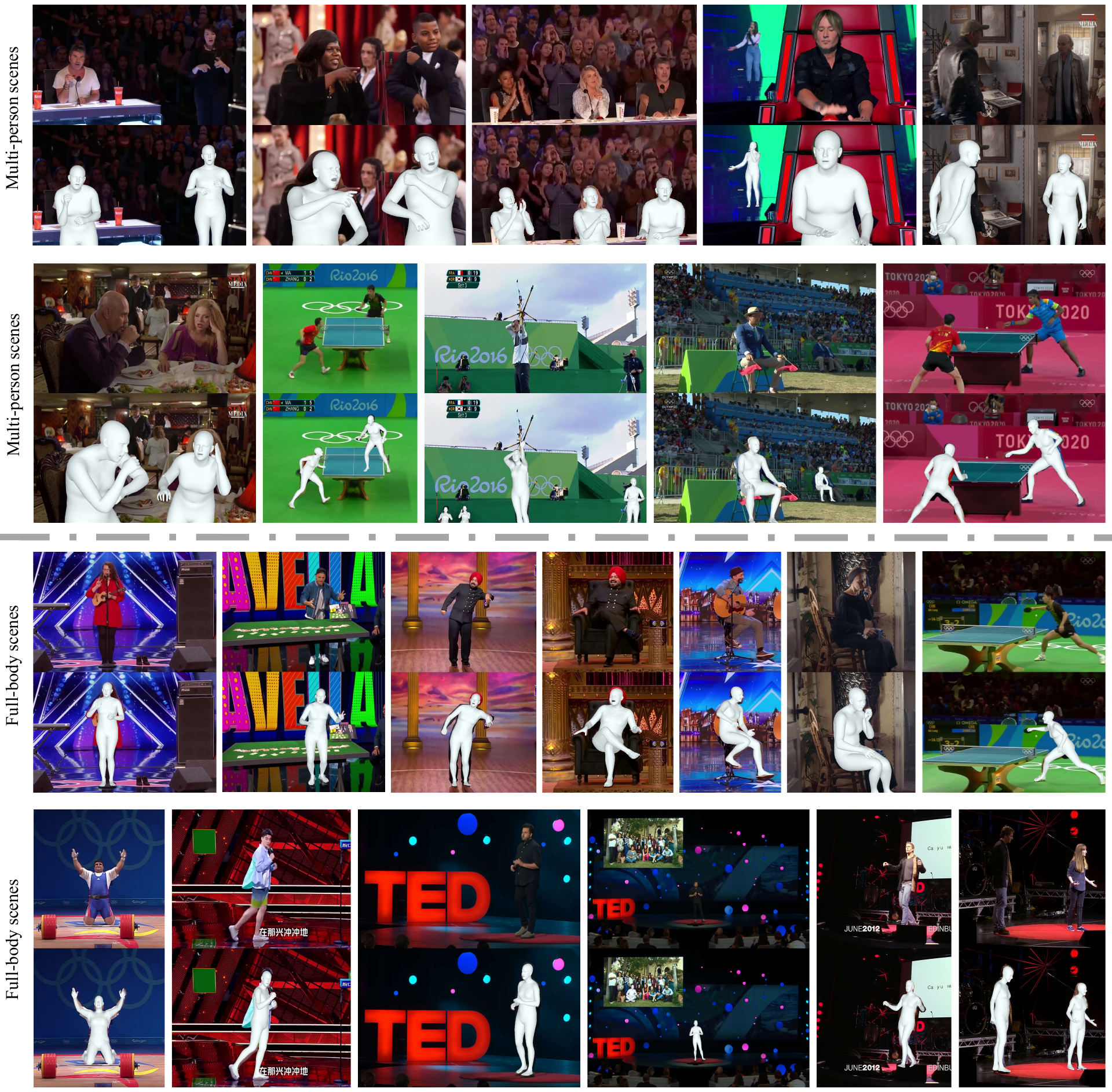}
\end{center}
\vspace{-0.6cm}
\caption{
Illustration of the ground-truth SMPL-X annotation for some special cases: \emph{multi-person scenes} and \emph{full body scenes} in \dataname. Our annotation pipeline can still work well on these scenes.
}
\label{fig:vis_ubody_gt6}
\end{figure*}

\noindent\textbf{Visualization of our annotation on UBody:} 
The visualizations of our SMPL-X annotation in our \dataname can be found in Figure~\ref{fig:vis_ubody_gt1}, \ref{fig:vis_ubody_gt2}, and \ref{fig:vis_ubody_gt6}. Our annotation produces high-quality ground truth. In many challenging cases of expressive hand poses, our estimated mesh can capture fine-level details.  

% \begin{figure*}[h]
% \begin{center}
% \includegraphics[width=1\linewidth]{fig/ubody_gt3.pdf}
% \end{center}
% \vspace{-0.6cm}
% \caption{
% Illustration of the ground-truth SMPL-X annotation for the three scenes in \dataname. For each scene, we show the input image (the upper) and our annotation (the lower).
% }
% \label{fig:vis_ubody_gt3}
% \end{figure*}

% \begin{figure*}[h]
% \begin{center}
% \includegraphics[width=1\linewidth]{fig/ubody_gt4.pdf}
% \end{center}
% \vspace{-0.6cm}
% \caption{
% Illustration of the ground-truth SMPL-X annotation for the three scenes in \dataname. For each scene, we show the input image (the upper) and our annotation (the lower).
% }
% \label{fig:vis_ubody_gt4}
% \end{figure*}

% \begin{figure*}[h]
% \begin{center}
% \includegraphics[width=1\linewidth]{fig/ubody_gt5.pdf}
% \end{center}
% \vspace{-0.6cm}
% \caption{
% Illustration of the ground-truth SMPL-X annotation for the three scenes in \dataname. For each scene, we show the input image (the upper) and our annotation (the lower).
% }
% \label{fig:vis_ubody_gt5}
% \end{figure*}

